\documentclass[letterpaper, 10 pt, journal, twoside]{IEEEtran}

\usepackage{graphicx}
\usepackage{amsmath}
\usepackage{algorithm}
\usepackage{algpseudocode}
\usepackage{xcolor}
\usepackage{soul}
\usepackage{stfloats}
\usepackage{amsfonts} 
\usepackage{amssymb}
\usepackage{tabularx}
\usepackage{booktabs}
\usepackage{comment}
\usepackage[caption=false]{subfig}

\usepackage{xcolor}
\usepackage{comment}
\usepackage{hyperref}
\newcommand\edit[1]{{\color{black}#1}}
\newcommand\ced[1]{{\color{black}#1}}
\newcommand\third[1]{{\color{black}#1}}
\usepackage{indentfirst}
\usepackage{todonotes}
\usepackage{tkz-graph}
\usetikzlibrary{automata,positioning}
\usepackage{tikz}
\usetikzlibrary{positioning}

\ifCLASSINFOpdf
\else
\fi

\begin{document}

\title{Log-GPIS-MOP: A Unified Representation for Mapping, Odometry and Planning}

% option1: Log-GPIS-MOP: Mapping, Odometry and Planning with A Unified Representation
% option2: MOP-Log-GPIS: Mapping, odometry and planning with a unified representation.
% option3: Log-GPIS: A Unified? Faithful? Representation for Odometry, Mapping and Planning
% option4: Log-GPIS-MOP: A Unified Representation for Mapping, Odometry and Planning (yes)

\author{Lan Wu, Ki Myung Brian Lee, Cedric Le Gentil, and Teresa Vidal-Calleja

\thanks{This work was supported by the Australian Government Research Training Program. All authors are with the UTS Robotics Institute, Faculty of Engineering and IT, University of Technology Sydney, NSW 2007, Australia. (Corresponding author: Lan Wu.) {\tt\footnotesize Lan.Wu-2@uts.edu.au}}
\thanks{This article and the supplementary downloadable material are available at https://ieeexplore.ieee.org, provided by the authors. The material consists of a video. This video does not aim to describe the method of the proposed approach, but it shows the experimental results of the proposed method.}
\thanks{© 2023 IEEE.  Personal use of this material is permitted.  Permission from IEEE must be obtained for all other uses, in any current or future media, including reprinting/republishing this material for advertising or promotional purposes, creating new collective works, for resale or redistribution to servers or lists, or reuse of any copyrighted component of this work in other works.}
}

% \markboth{IEEE TRANSACTIONS ON ROBOTICS}%
% {Wu \MakeLowercase{\textit{et al.}}: Log-GPIS-MOP: A Unified Representation for Mapping, Odometry and Planning}

% \IEEEpubid{0000--0000/00\$00.00~\copyright~2023 IEEE}

\maketitle

\begin{abstract}
Whereas dedicated scene representations are required for each different task in conventional robotic systems,
this paper demonstrates that a unified representation can be used directly for multiple key tasks. We propose the Log-Gaussian Process Implicit Surface for Mapping, Odometry and Planning (Log-GPIS-MOP): a probabilistic framework for surface reconstruction, localisation and navigation based on a unified representation. Our framework applies a logarithmic transformation to a Gaussian Process Implicit Surface (GPIS) formulation to recover a global representation that accurately captures the Euclidean distance field with gradients and, at the same time, the implicit surface. By directly \edit{estimating} the distance field and its gradient through Log-GPIS inference, the proposed incremental odometry technique computes the optimal alignment of an incoming frame and fuses it globally to produce a map. Concurrently, an optimisation-based planner computes a safe collision-free path using the same Log-GPIS surface representation. We validate the proposed framework on simulated and real datasets in 2D and 3D and benchmark against the state-of-the-art approaches. Our experiments show that Log-GPIS-MOP produces competitive results in sequential odometry, surface mapping and obstacle avoidance.
\end{abstract}

\begin{IEEEkeywords}
Gaussian Process Implicit Surfaces, Euclidean Distance Field, Mapping, Localisation, Planning, SLAM.
\end{IEEEkeywords}

\IEEEpeerreviewmaketitle

\section{Introduction}\label{introduction section}
% key information: representation is important for localisation, mapping and path planning.
%TCV: This can easily go to the introduction
\IEEEPARstart{S}{imultaneous} localisation and mapping (SLAM) is a well-known problem in robotics~\cite{Cadena16tro-SLAMfuture}, which concerns reconstructing or building a map of an unknown environment while simultaneously tracking the location of a robot-sensor system within the map. 
Using the built map and the estimated location, a path planner may produce collision-free paths for the robot to navigate safely. 
Each different task raises varying requirements leading to dedicated environment representations. 
The choice of scene representation strongly affects the performance of the localisation, mapping, and path planning tasks.

\begin{figure}[t]
  	\centering
	\resizebox{1\linewidth}{!}{
	\includegraphics[]{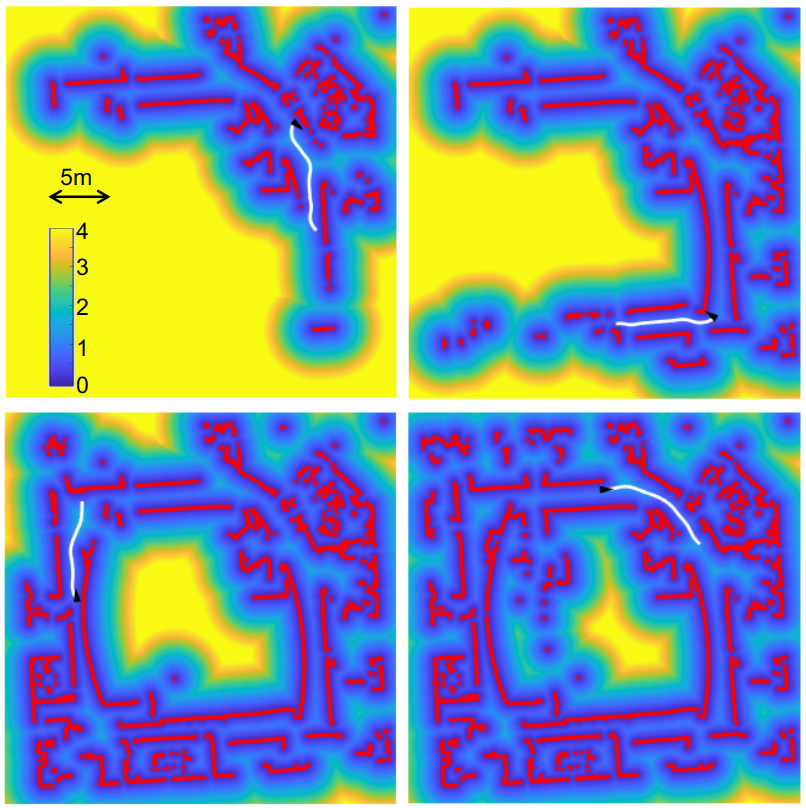}
	}
	\caption{The proposed framework Log-GPIS-MOP incrementally builds the distance field (yellow to blue colour map) and estimates the implicit surface (red) of the first loop around the Intel Lab~\cite{intel}. Our incremental odometry and planning use the distance and gradients of the Log-GPIS as input to compute the current alignment and the white optimal trajectory for the robot to avoid colliding with the surfaces respectively.}
    \vspace{-2ex}
	\label{fig:teaser}
\end{figure}

For instance, motion estimation favours sparse representations such as the locations of features in the environment, which allows consistent estimation of the robot's location. 
A key objective of the mapping task is to reconstruct an accurate, dense and high-resolution map of the scene for, e.g., inspection purposes. 
Path planning similarly requires dense information such as obstacle occupancy or closest distance to collision in order to avoid obstacles. 
% \IEEEpubidadjcol
As a unified representation for multiple purposes, a feature-based representation such as a set of landmarks\edit{~\cite{paper:orbslam2,thoma2019mapping}} is able to reconstruct a sparse map along with self-localisation and path planning in an unknown environment.
Although sparse feature-based representations have been widely used for representing geometric structures and have proven advantageous for localisation\edit{~\cite{paper:orbslam2,vins-mono}}, they are not suitable for an accurate depiction of the environment or path planning, where dense maps are more appropriate, although difficult to recover from sparse features. As an alternative, several approaches~\cite{kinectfusion,Chomp} employ signed distance functions (SDF) as a representation for path planning and dense mapping. 
SDF can be used directly by a planner and the implicit surfaces can be used to recover the dense map. 
However, without matching sparse features to identify correspondences, it is difficult to perform a consistent alignment for motion estimation for localisation using the SDF representation~\cite{reijgwart2020voxgraph}.

In this work, we propose a unified representation so-called Log-GPIS suitable for localisation, mapping, and path planning (see Fig.~\ref{fig:teaser}), inspired by recent work on SDF representations with Gaussian process implicit surfaces (GPIS). GPIS~\cite{Microsoft},~\cite{LanRAL20},~\cite{paper:GeomPrior} is a continuous and probabilistic representation that exploits the implicit surfaces. 
Whereas conventional GPIS representations~\cite{Bhoram,stork2020ensemble} can only infer accurately the Euclidean distance field (EDF) close to the measurements (noisy points on the surface), Log-GPIS, proposed in our previous work~\cite{wu2021faithful}, faithfully models the EDF and gradients by enforcing an approximation of the Eikonal equation~\cite{Crane} through a simple log transformation on the standard GPIS.
This representation allows us to use Log-GPIS as a single unified representation for localisation, mapping and planning multiple robotics tasks, which is the focus of this paper. The smooth and probabilistic nature of Log-GPIS provides an accurate dense surface for mapping given noisy measurements. The extrapolative power of Log-GPIS provides information not only near but also far away from the measurements, and hence enables estimation of the full EDF with gradients. 
This allows motion estimation via direct optimisation of the alignment between the global map representation and a new set of measurements. 
Furthermore, the EDF and its gradients are suitable for trajectory optimisation-based planners. 
% We name the proposed framework Log-GPIS for mapping, odometry, and planning (Log-GPIS-MOP). 

The contributions of this paper are as follows:

\begin{itemize}
\item The Log-GPIS-MOP framework, which achieves incremental localisation, mapping and planning based on a single continuous and probabilistic representation through accurate prediction of the EDF and the implicit surface with gradients.

\item A sound formulation for 2D/3D point cloud alignment for odometry estimation, where the Log-GPIS is used to minimise the point cloud distance to the implicit surface from a set of newly arrived observations following the gradients.

\item A generic mapping approach for organised and unorganised point clouds that consists in probabilistically fusing new measurements into a global Log-GPIS representation and is ready for a modified version of marching cubes to recover dense surface reconstruction. 

\item A path planning approach that uses Log-GPIS to avoid collisions with the mapped surfaces following the EDF and gradient information. 

%\item 
\end{itemize}

Moreover, we present a thorough evaluation of the overall Log-GPIS-MOP framework and individual tasks on public benchmarks, and qualitative and quantitative benchmarking with the state-of-the-art. Our work enables sequential and concurrent localisation, mapping and planning, focusing on accurate performance as opposed to real-time operations.

% key information: paper structure
The paper structure is organised as follows. The related work is presented in Sec.~\ref{related work section}. Sec.~\ref{background section} explains the background knowledge. Our Log-GPIS representation is discussed in Sec.~\ref{log gpis section}. The methodology of the Log-GPIS-MOP framework is presented in Sec.~\ref{method overview}~, with odometry in~\ref{odometry section}, mapping~\ref{mapping section}, and planning~\ref{planning section}. In Sec.~\ref{results section}, we present a thorough evaluation of the framework on public benchmarks and compare the performance to existing solutions. Finally, the conclusion and suggestions on future work are given in Sec.~\ref{conclusion section}.

\section{Related Work}\label{related work section}
% key information: talk about feature-based slam representation.
For visual SLAM, several successful frameworks process the input data into a set of sparse features as the representation for odometry and mapping. Recent feature-based SLAM systems~\cite{paper:orbslam2, schneider2018maplab, vins-mono} have demonstrated accurate pose estimation, consistency throughout long-term operations, and real-time performance. Due to the difficulties of extracting continuous surfaces from a sparse set of points, most of them are not suitable for applications other than localisation. In addition, dedicated and sophisticated algorithms are required to recover a dense and accurate reconstruction from sparse points~\cite{MVE}.

% key information: talk about KinectFusion.
As an alternative representation to achieve sensor tracking and mapping, the seminal work of KinectFusion~\cite{kinectfusion} demonstrates high-performance camera motion estimation and dense reconstruction in real-time. To represent the geometry, KinectFusion uses the truncated signed distance function (TSDF)~\cite{tsdf} and a coarse-to-fine ICP algorithm~\cite{ICP} to perform an alignment. TSDF, originally proposed in computer graphics literature, has become a widely used representation in odometry and mapping approaches. However, KinectFusion has some drawbacks. First, the algorithm implementation relies on GPU. Second, the distance of the TSDF is truncated and generally overestimates the distance within limited viewpoints. Third, the representation can not be directly used for planning.

% key information: RGB-D slam.
Another relevant representation for mapping and odometry tracking is from RGB-D SLAM~\cite{rgbd-mapping}. Such as in the framework recently proposed in~\cite{Whelan2015ElasticFusionDS}, ElasticFusion represents the map as a collection of surfels, which are small surface elements with normals and colours. Given a surfel-based representation, ElasticFusion incrementally estimates the sensor pose through dense frame-to-map tracking. No pose-graph optimisation or post-processing is required. Such representation generally assists in computing accurate pose estimation, along with high-quality reconstruction and a rich understanding of the occupied space. However, the representation does not differentiate free space from unknown space explicitly, which is of limited applicability for autonomous navigation tasks. 

% talk about the semantic slam and planning as the reviewer asked
\edit{Semantic combined with geometric representations are a recent development gaining popularity for mapping~\cite{zobeidi2020semanticGP} and localisation~\cite{rosinol2020kimera}. The application has since been further extended for navigation~\cite{vasilopoulos2020semantic_planning}. It provides competitive efficiency, accuracy and robustness for metric-semantic SLAM and perception. However, the mesh representation used in these works does not provide inherently distance and direction to the collision which has to be computed separately for navigation.}%there is no explicit occupancy or distance information immediately available for rich navigation and path planning. } 

% key information: occupancy grid mapping
\edit{
Other representations based on explicit occupancy~\cite{occupancy} efficiently compute and store information, such as free space, occupied space, and unknown. Occupancy maps are commonly used for mapping and path-planning applications. Many options for occupancy maps have been presented in the literature varying from discrete representations~\cite{occupancy}~\cite{Octomap}~\cite{carto2016} to probabilistic and continuous ones~\cite{occupancy2012gaussian,OccMaps}. Discrete occupancy maps comprise a set of grid cells with occupancy information regarding the environment. Planning algorithms frequently take such discrete occupancy grids as input~\cite{occupancy2016review}. As for the probabilistic and continuous options, the authors in~\cite{OccMaps2} propose continuous occupancy mapping by viewing Gaussian processes as implicit functions of occupancy. Doing so, it is capable of building the representation with both occupancy information and implicit surfaces. Extension of occupancy grids from 2D to 3D is, however, challenging, due to large memory usage. It is often difficult to perform fast ray-casting and look-up tabling for any space larger than a room. To this end, an online solution commonly used in 3D is Octomap~\cite{Octomap}. This approach uses an octree-based structure to represent the occupancy probabilities of cells. The octree structure allows handling large spaces by immensely decreasing the amount of memory required to represent unknown or free space. 
}

% key information: occupancy is not enough for some planners.
\edit{
However, the representation with occupancy information alone is often insufficient for planning approaches. For instance, trajectory optimisation-based planners including CHOMP~\cite{Chomp} and TrajOpt~\cite{TrajOpt}, require information on distance and direction to obstacles. In addition, having a distance map speeds up collision checking of complex environments. For example, robot arms are commonly represented as a set of overlapping spheres, in which case, it is much more convenient to query the distance field at the centre of each sphere for collision checking~\cite{Chomp}.
}

\edit{
% the edf and gradient help the planner with details
Such optimisation-based planners require a complete distance field that is not truncated further from the surface. For this reason, a \third{Euclidean signed distance field (ESDF)} containing distance values over the entire space is necessary. An ESDF can be naturally passed to a collision-checking algorithm to produce collision-free trajectories~\cite{Chomp}~\cite{TrajOpt}. Gradients can be computed using the distance values of the neighbouring cells, which allows CHOMP and other planners to follow the gradient direction of the distance to push the trajectory away from collision~\cite{Helen2016signed}. 
}

% voxblox is introduced here.
\edit{
Following the above requirements, Oleynikova \emph{et al}~\cite{Voxblox} proposed the Voxblox framework to incrementally generate a representation for mapping and planning. It adopts the advantages of TSDF to generate an implicit surface, as its denseness is ideal for human visualisation. When using TSDF, the distance within the truncated threshold is relatively accurate. Then, an ESDF representation is numerically approximated from the TSDF to cover the observations' fields of view. The basic unit for ESDF is a voxel grid, which contains the distance to the nearest obstacles and the gradient thereof. The ESDF can be used with trajectory optimisation methods for online navigation, such as the one in~\cite{oleynikova2018loco} proposed by the same authors. \edit{Similarly, FIESTA~\cite{han2019fiesta} proposes a discrete voxel grid representation with SDF for mapping and planning, which reported comparable results to~\cite{ortiz2022isdf}.}
}

% talk about voxgraph here.
Compared to KinectFusion, Voxblox does not use the distance information from the TSDF or ESDF to perform motion estimation. The system still requires ground truth poses as input from a motion-capturing system or estimated poses from a visual-inertial odometry framework~\cite{rovio}. To make the distance play a role in the trajectory estimation, the same authors extended their work to tackle the long-term consistent mapping problem in Voxgraph~\cite{reijgwart2020voxgraph}. It introduces submap-based ESDF and TSDF representations for mapping and navigation. Voxgraph adopts the distance information to generate correspondence-free constraints between submaps. \edit{Then, through optimising a pose graph, the relative pose given submaps in a closed loop is computed to mitigate long-term drifts. However, Voxgraph still requires external visual odometry to produce submaps from the incoming sensor data for further optimisation and localisation.}

\edit{An interesting recent development is the use of neural networks for continuous implicit representation for reconstruction and mapping~\cite{gropp2020learning_shapes,park2019deepsdf,mildenhall2021nerf,mescheder2019occupancy_networks}. Neural representations can be trained to model radiance fields~\cite{mildenhall2021nerf}, signed distance functions~\cite{gropp2020learning_shapes,park2019deepsdf,sitzmann2020implicit} or occupancy maps~\cite{mescheder2019occupancy_networks}. Most prominent is the neural radiance field~(NeRF)~\cite{mildenhall2021nerf}, which represents the 3D environment with a continuous volumetric density function in tandem with view-dependent colours, both modelled by a neural network. 
Such neural representations have shown great promise for SLAM and localisation~\cite{sucar2021imap,rosinol2022nerf_slam,zhu2022nice_slam} as well as path planning~\cite{ortiz2022isdf,driess2022learningSDF_planning,adamkiewicz2022planning_using_NeRFs,pantic2022NeRF_planning}. iMAP~\cite{sucar2021imap} first proposed to embed NeRF within a SLAM framework, followed by NICE-SLAM~\cite{zhu2022nice_slam}, which uses multiple NeRFs organised into a voxel grid.
Using the constructed NeRF, trajectory planners have been proposed to optimise the path to avoid obstacles, which can be identified in NeRFs as regions of high density~\cite{adamkiewicz2022planning_using_NeRFs,pantic2022NeRF_planning}.
Of those, \cite{pantic2022NeRF_planning} shows that planning performance can be improved by using an approximate ESDF derived from a NeRF.
Our work learns an EDF, an unsigned variant of ESDF, which facilitates all of localisation, mapping and planning. 
Further, although these neural representations also serve as a sound representation for multiple tasks, we achieve similar functionalities with GPs, which we believe is significantly more `white-box' than a neural network, and hence easier to understand and diagnose for practitioners.
This comes with the added benefit of explicit uncertainty representation, which is not readily available with neural network models. 

Furthermore, learning SDFs directly with neural network models has also gained popularity in recent years~\cite{gropp2020learning_shapes,park2019deepsdf,sitzmann2020implicit}, which has been also extended to mapping and navigation~\cite{ortiz2022isdf}.
A recent work closely related to ours is \cite{gropp2020learning_shapes}, which reports improvement in model performance through the use of a penalty to account for the Eikonal equation.
The penalty is incorporated into the loss function, and the Eikonal equation is thus enforced via multiple gradient updates.
In contrast, our framework elegantly incorporates a regularised version of the Eikonal equation through a single log transformation to accurately approximate the true EDF.

% There have been works to explore these types of representations for SLAM and localisation~\cite{sucar2021imap,rosinol2022nerf_slam,zhu2022nice_slam}. Through extensive work, the continuous implicit neural-based representation is suitable for multiple purposes. As an alternative solution and one step further, our proposed representation is continuous, implicit, and also probabilistic to deal with uncertainty.
}

% % other unified representation for multi-tasks
% The authors proposed a landmark-based representation in~\cite{thoma2019mapping} for sparse map construction, self-localisation and path planning. The representation is a set of image landmarks selected from a sequence of images. The localisation is implemented by computing a flow between the landmarks, while at the same time, the shortest path is computed along the selected landmarks. However, sparse landmarks represent geometric structures properly, but they are not suitable for dense reconstruction or path planning due to their sparseness and discontinuity. In contrast, our paper uses a single and unified representation generated from depth data only for odometry estimation, continuous dense mapping and planning with no aid from any other positioning system.

% shall we talk about gpis and log-gpis here?
As mentioned above our representation is based on Gaussian Process implicit surfaces (GPIS), a continuous and probabilistic representation that was initially proposed to reconstruct surfaces given noisy surface observations~\cite{Microsoft}. Later in~\cite{Bhoram} \edit{and~\cite{lee2019thesis}}, the authors used the GPIS as an approximate signed distance function (GPIS-SDF) that infers the distance to the nearest surfaces. \edit{Based on the distance values, a localisation solution is formulated to find consecutive poses in~\cite{lee2019thesis}.} However, it only keeps track of the distance information close to the surface as there is a lack of observations further away. \edit{The localisation requires the Kalman filter to provide the initial guess and odometry is performed with frame-to-map registration but does not leverage a pose graph post-processing step as proposed in here to further improve the estimates' accuracy.} Based on the Varadhan's approach~\cite{Varadhan}, our previous work in~\cite{wu2021faithful} applies the logarithmic transformation to a GPIS with a specific covariance function to recover the distance field in the space. This method provides the extra potential for GPIS to accomplish accurate distance field mapping even further away from the surface. However, our previous work does not consider solving the localisation problem as it assumes the sensor poses are known and do not actually propose a planning approach that leverages such a representation. %The paper here generalises the use of Log-GPIS in a simultaneous and incremental mapping

% our representation is for three purposes.
Through the related works, catering for the various requirements to have a representation for multiple purposes is still an open research problem. Thus, in the paper, we present the unified environment representation that can be used for dense reconstruction, collision avoidance, while providing additional information needed for the sensor's odometry estimation. 

\section{Background}\label{background section}

In this section, we introduce the Eikonal equation, the derivation of the distance approximation used in this work and a brief description of the Gaussian Process Implicit Surfaces.

\subsection{Euclidean Distance Field}
\edit{We assume that a robot's given environment can be modelled as} a bounded \edit{manifold} denoted as $S \subset \mathbb{R}^{D}$ with a suitably smooth boundary $\partial S$. 
The boundary $\partial S$ \edit{is assumed to be orientable, which means it has a continuously varying surface normal at every point. 
We seek to represent the boundary $\partial S$ in terms of its EDF $d(\mathbf{x})$, which is the nearest distance between a given point $\mathbf{x} \in \mathbb{R}^{D}$ and points on the boundary $\mathbf{w} \in \partial S$:}
\begin{equation}\label{eq:edf}
    d(\mathbf{x})=\min_{\mathbf{w} \in \partial S}  | \mathbf{x} - \mathbf{w} | \,.
\end{equation}
\edit{The EDF~\eqref{eq:edf} represents the boundary $\partial S$ implicitly as its zero-level set.}

\edit{A definitive property of EDFs is that it satisfies the \emph{Eikonal equation} \third{almost everywhere\footnote{\third{Unsigned EDFs are not differentiable at the boundary $\partial S$, and hence the Eikonal equation does not hold at $\partial S$.}}, which states that the gradient is of unit norm:}
\begin{equation}\label{eikonal}
    |\nabla d(\mathbf{x})|=1.
\end{equation}
The intuition behind the Eikonal equation is that the fastest increase in the EDF $d(\mathbf{x})$ with a unit change in $\mathbf{x}$ is a unit change in distance, which occurs when moving in the normal direction to the surface.
Since the gradient $\nabla d(\mathbf{x})$ represents the direction and rate of fastest increase in $d(\mathbf{x})$, it has a unit magnitude and is directed normally to the surface.
Our framework represents the EDF \emph{faithfully} by closely approximating the Eikonal equation~\eqref{eikonal} through a simple log transformation.
}

\subsection{Heat-based distance function}
\label{sec:heat_dist}
% foremost concern
In the robotics literature, recently proposed methods such as~\cite{Voxblox}~\cite{lau2010improved} adopted the concept of wavefront propagation to estimate the distance field efficiently. This approach incrementally propagates the distance field to the neighbours directly from TSDF or from the occupancy map through a voxel grid. In this way, a compromise has to be made to not obtain the EDF directly as further post-processing algorithm is required. To recover the distance field exactly, one option is to solve \edit{or approximate} the Eikonal equation. However, the main restriction in exploiting the Eikonal equation~\eqref{eikonal} is that the equation is hyperbolic and non-linear, which makes it difficult to directly solve and recover the accurate distance values.  %implement fast marching or label-correcting to propagate the distance field 

\ced{
Inspired by the idea proposed in the computer graphics literature~\cite{Crane}, we motivate our work based on the physical phenomenon of heat propagation in space.
The main idea is that the amount of heat propagated from a heated surface for infinitesimal time reflects the closest distance to the surface.

Let us consider the surface $\partial S$ to be heated at an arbitrary temperature of one. 
Formally, the heat equation, the boundary conditions, and initial conditions are as follows:
\begin{equation}
    \frac{\partial v(\mathbf{x},t)}{\partial t} = \Delta v(\mathbf{x},t) \ \text{with}\begin{cases}
  v(\mathbf{x},t) = 1\ \text{when}\ \mathbf{x}\in \partial S \\    
  v(\mathbf{x},0) = 0\ \text{when}\ \mathbf{x}\notin \partial S,
\end{cases}
    \label{eq:heat_eq}
\end{equation}
where $\Delta$ is the Laplacian $\sum_{i=0}^{D} \frac{\partial^2}{\partial x_i^2}$, and $v(\mathbf{x},t)$ is the heat at location $\mathbf{x}\in \mathbb{R}^D$ and time $t$.
%Rewriting the temporal differentiation with limits around time $t=0$ and leveraging the initial conditions\footnote{Using the fact that $\frac{\partial v(\mathbf{x},t)}{\partial t}|_{t=0} = \lim_{t\to 0}\frac{v(\mathbf{x},t)\text{-}v(\mathbf{x},0)}{t}$ and $v(\mathbf{x},0) = 0$ leads to $\frac{\partial v(\mathbf{x},t)}{\partial t}|_{t=0} = \lim_{t\to 0}\frac{v(\mathbf{x},t)}{t}$}, \eqref{eq:heat_eq} becomes 
For a small time period around $t=0$, the heat equation~\eqref{eq:heat_eq} can be approximated as a screened Poisson equation\footnote{\ced{This is by discretising the temporal differentiation using the fact that $\frac{\partial v(\mathbf{x},t)}{\partial t}|_{t=0} = \lim_{t\to 0}\frac{v(\mathbf{x},t)\text{-}v(\mathbf{x},0)}{t}$ and $v(\mathbf{x},0) = 0$, which leads to $\frac{\partial v(\mathbf{x},t)}{\partial t}|_{t=0} = \lim_{t\to 0}\frac{v(\mathbf{x},t)}{t}$.}}:
\begin{equation}
%    \lim_{t\to 0}\frac{v(\mathbf{x},t)}{t} = \Delta v(\mathbf{x},t).
    v(\mathbf{x},t) = t \Delta v(\mathbf{x},t).
    \label{eq:lim_heat_eq}
\end{equation}
The celebrated result of Varadhan \cite[Theorem~2.3]{Varadhan} is that the solution to~\eqref{eq:lim_heat_eq} is related to the EDF as follows:\footnote{\ced{Varadhan's original theorem only applies to points in $S$~\cite{Varadhan}, but we proposed to extend it to all points in $\mathbb{R}^{D}$ by applying the theorem again to the complement $S^{c}$ (i.e. outside $S$)~\cite{wu2021faithful}.}
%Note that our method does not require a closed surface as Varadhan's result generalises to any shaped surface.
\label{fn:domain_extension}
}}
\begin{equation}
    d(\mathbf{x}) = \lim_{t\to 0} \{-\sqrt{t}\ln\left(v(\mathbf{x}, t)\right)\}.
    \label{eq:dist_approx}
\end{equation}
\ced{Intuitively, \eqref{eq:dist_approx} states that the negative logarithm of the heat propagated from a surface after a small quantum of time is proportional to the distance to the surface.

Importantly, Varadhan's formula~\eqref{eq:dist_approx} enables approximating the EDF as $d(\mathbf{x})\approx-\sqrt{t}\ln\left(v(\mathbf{x}, t)\right)$ for a sufficiently small $t$.
The Eikonal equation~\eqref{eikonal} remains approximately satisfied by this approximate EDF~\cite{Belyaev}.
Namely, substituting $v(\mathbf{x},t) \approx \exp\left(-\frac{d(\mathbf{x})}{\sqrt{t}}\right)$ into \eqref{eq:lim_heat_eq} recovers a regularised form of Eikonal equation~\eqref{eikonal}: 
\begin{align}
%    \frac{v(\mathbf{x},t)}{t} & = \frac{v(\mathbf{x},t)}{t}\left(\lvert\nabla d(\mathbf{x})\rvert^2 - \frac{\Delta d(\mathbf{x})}{\sqrt{t}} \right) \\
%    \implies
    1 & = \lvert\nabla d(\mathbf{x})\rvert^2 - \sqrt{t}\Delta d(\mathbf{x}),
    \label{eq:regularised_eikonal}
\end{align}
where $\sqrt{t}\Delta d(\mathbf{x})$ is the regulariser.
Accordingly, it is clear that the heat-based approximation \eqref{eq:dist_approx} improves as the time $t$ decreases.
}
\subsection{Gaussian Process Implicit Surfaces}
Gaussian Process (GP)~\cite{GPbook} is a non-parametric stochastic method ideal for solving non-linear regression problems. GPIS techniques~\cite{Microsoft},~\cite{LanRAL20},~\cite{paper:GeomPrior},\edit{~\cite{liu2021active}} use a GP approach to estimate a probabilistic and continuous representation of the implicit surface given noisy measurements. \edit{Further proposed in~\cite{GPmap,Bhoram,stork2020ensemble}}, GPIS representation can be used for distance field estimation, which provides approximate distance values only near the surface. 

\edit{The following provides the background knowledge of GPIS formulation for implicit surface and distance estimation. GP is capable of modelling the unknown function with gradients because the derivative of a GP is another GP~\cite{GPbook,paper:GeomPrior}. In the GPIS formulation, considering the unknown function, $d$ represents the distance to a surface and $\nabla d$ denotes the corresponding gradient of $d$. Then $d$ with $\nabla d$ can be modelled by the joint GP with zero mean:}
\begin{equation}
    \begin{bmatrix}
        d \\
        \nabla d
    \end{bmatrix} \sim \mathcal{GP}({0}, \tilde{k}(\mathbf{x}, \mathbf{x}')).
\end{equation}
and covariance matrix
% \begin{equation}
%     \tilde{\edit{k}}\left(\mathbf{x}, \mathbf{x}'\right)=\left[\begin{array}{cc}
% k\left(\mathbf{x}, \mathbf{x}'\right) & k\left(\mathbf{x}, \mathbf{x}'\right) \nabla_{\mathbf{x}'}^{\top}  \\
% \nabla_{\mathbf{x}} k\left(\mathbf{x}, \mathbf{x}'\right) & \nabla_{\mathbf{x}} k\left(\mathbf{x}, \mathbf{x}'\right) \nabla_{\mathbf{x}'}^{\top}
% \end{array}\right],
% \end{equation}
\begin{equation}
    \tilde{k}\left(\mathbf{x}, \mathbf{x}'\right)=\left[\begin{array}{cc}
k\left(\mathbf{x}, \mathbf{x}'\right) & k\left(\mathbf{x}, \mathbf{x}'\right) \nabla_{\mathbf{x}'}^{\top}  \\
\nabla_{\mathbf{x}} k\left(\mathbf{x}, \mathbf{x}'\right) & \nabla_{\mathbf{x}} k\left(\mathbf{x}, \mathbf{x}'\right) \nabla_{\mathbf{x}'}^{\top}
\end{array}\right],
\end{equation}
where $k(\mathbf{x}, \mathbf{x}')$ is the covariance function of the function $d$. $\nabla_{\mathbf{x}}$ and $\nabla_{\mathbf{x}'}$ are the partial derivatives of $k(\mathbf{x}, \mathbf{x}')$ at $\mathbf{x}$ and $\mathbf{x}'$\edit{~\cite{paper:GeomPrior}~\cite{GPbook}}. 

Let us consider a set of \edit{points and the measurements of} points corrupted by noise. The distance measurements $\mathbf{y}=\left\{y_{j} = \edit{d}(\mathbf{x}_{j}) + \epsilon_{j} \right\}_{j=\edit{0}}^{J} \subset \mathbb{R}$ and the corresponding gradients $\nabla \mathbf{y} \subset \mathbb{R}^{D}$ are taken at locations $\mathbf{x}_{j} \in \mathbb{R}^{D}$ and affected by additive Gaussian noise $\epsilon_{j} \sim \mathcal{N}(0, \sigma_{\edit{d}}^2)$. \edit{The set of point positions, the measurements and the corresponding gradients are the input training data.} The posterior distribution of $d$ at an arbitrary testing point $\mathbf{x}_{*}$ is given by $d(\mathbf{x}_{*}) \sim \mathcal{N}(\bar{d}_{*},\mathbb{V}\left[d_{*}\right])$ with the predictive mean and variance with derivatives are expressed as follows:
\begin{equation}\label{eq:gp_inference_gradients}
\begin{aligned}
\begin{bmatrix}
\bar{d}_{*} \\
\nabla \bar{d}_{*}
\end{bmatrix} &=
 \tilde{\mathbf{k}}_{*}^{\top}(\tilde{K}+\sigma_{d}^2 I)^{-1}\begin{bmatrix}
\mathbf{y} \\
\nabla \mathbf{y}
\end{bmatrix}
\end{aligned}
\end{equation}
\begin{equation}\label{eq:gp_inference_gradients_var}
\begin{aligned}
\begin{bmatrix}
\mathbb{V}\left[d_{*}\right] \\
\mathbb{V}\left[\nabla{d}_{*}\right]
\end{bmatrix} &=
 \tilde{k}(\mathbf{x}_{*}, \mathbf{x}_{*})-
 \tilde{\mathbf{k}}_{*}^{\top}(\tilde{K}+\sigma_{d}^2 I)^{-1}\tilde{\mathbf{k}}_{*}.
\end{aligned}
\end{equation}
Note that $I$ represents the identity matrix of size $J(D+1) \times J(D+1)$. $\tilde{\mathbf{k}}_{*}$ represents the covariance vector between the training points and the testing point $\mathbf{x}_{*}$. $\tilde{K}$ is the covariance matrix of the training points $J$ with gradients. Furthermore, $\tilde{k}(\mathbf{x}_{*}, \mathbf{x}_{*})$ is the covariance function between the testing point.

Most GPIS techniques ensure the accuracy of the reconstructed surfaces, but only a few of them directly model the distance field~\cite{Bhoram,stork2020ensemble}. However, one of the disadvantages of standard GPIS applying inference of the distance field is that only accurately infers the distance near the surface. Given the limited training data, it is hard to keep the high precision and continuity of the distance field further from the measurements. \edit{To overcome this shortcoming, we introduce the Log-GPIS representation in Sec.~\ref{log gpis section}.}
\ced{Based on the physics of heat propagation, Log-GPIS leverages both the well-known GPIS formulation and Varadhan's results to approximate the Euclidean distance field over the full space $\mathbb{R}^D$ given minimal and sparse observations close to the surface.}
%It uses the log transformation of the well-known GPIS model after applying Varadhan's equation to predict the exact Euclidean distance field given minimal and sparse observations close to the surface.
With accurate distance predicted, we unlock the potential of Log-GPIS to accomplish \edit{multiple robotic} tasks. \edit{Other than mapping in Sec.~\ref{mapping section}, we will discuss the} Log-GPIS based odometry in Sec.~\ref{odometry section} and path planning in Sec.~\ref{planning section}.

\section{Log-Gaussian Process Implicit Surfaces}\label{log gpis section}
\third{In this section, we introduce the derivation of the Log-GPIS representation and the choice of covariance function suitable for Log-GPIS with gradient inferences.}
\subsection{Derivation}\label{sec:derivation}
\ced{Our Log-GPIS representation is built upon the heat-based distance approximation presented in Section~\ref{sec:heat_dist}.
We use a GP to represent the heat $v(\mathbf{x})$ and its gradient $\nabla v(\mathbf{x})$ after a quantum of time $t$:
}
\begin{equation}
    \begin{bmatrix}
        v \\
        \nabla v
    \end{bmatrix} \sim \mathcal{GP}({0}, \tilde{k}(\mathbf{x}, \mathbf{x}'))\,.
\end{equation}
\ced{The function $v(\mathbf{x})$ must be a solution of the screened Poisson PDE \eqref{eq:lim_heat_eq}.
This solution can be enforced by choosing a specific kernel function $\tilde{k}$.
We discuss the kernel choice in the following subsection.
}

Assuming that the choice of the covariance function \edit{respects the screened Poisson equation}, we estimate $\bar{v}(\mathbf{x})$, $\nabla \bar{v}$ and the corresponding variances by simply using the regression equations~\eqref{eq:gp_inference_gradients} and~\eqref{eq:gp_inference_gradients_var}.
\ced{By applying~\eqref{eq:dist_approx}, we recover the distance field $d(\mathbf{x})$  as:}
\begin{equation}
    \bar{d}_{*}=-\sqrt{t} \ln \bar{v}_{*}.
    \label{3}
\end{equation}
\edit{This logarithm transformation gave its name to our method, \emph{Log}-GPIS.
To ensure that $\bar{v}_{*}$ is positive, we use its absolute value in our implementation.}
\ced{The set of surface observations $\mathbf{y}$ in~\eqref{eq:gp_inference_gradients} corresponds to virtual heat observations on the surface~$S$.
In other words, the vector $\mathbf{y}$ is equal to $\mathbf{1}$ (cf. boundary conditions in~\eqref{eq:heat_eq}).}
\edit{We also use the surface normals, equivalent to the normalised gradient of the implicit distance function to compute the gradient observations $\nabla\mathbf{y}$.
}

Despite its simplicity, the log transformation has a non-trivial benefit as the predicted distance will approach infinity when querying points further away from the surface unlike the standard GPIS, which will predict a distance value of zero (falling back to the mean far away from the measurements).
In other words, Log-GPIS \edit{avoids} the undesirable artifacts further away from the predicted surface.
\ced{One shortcoming of the proposed method is that it does not predict} whether $\mathbf{x}$ is inside or outside the surface while \ced{standard GPIS does (i.e., Log-GPIS predicts EDF instead of SDF).
\footnote{
\ced{This follows from the extension of Varadhan's formula~\eqref{eq:dist_approx} from $S$ to $\mathbb{R}^{D}$ (see footnote \ref{fn:domain_extension}). Applying Varadhan's formula on both $S$ and the complement $S^{c}$ implies `stitching' the EDFs of $S$ and $S^{c}$, and hence whether a point belongs to $S$ is no longer distinguishable.
}}
}
In our perspective, given the advantage of using~\eqref{eq:lim_heat_eq} to approximate the distance field everywhere on $\mathbb{R}^{D}$, losing the sign is a small price to pay for the convenience of the simplicity and accuracy of the proposed distance estimation.

We also observe that the log transformation affects the gradient.
Taking the partial derivative of \eqref{3} with respect to \ced{the query point~$\mathbf{x}_*$} leads to:
\begin{equation}\label{eq:gradient}
    \nabla \bar{d}_{*} = \frac{-\sqrt{t}}{\bar{v}_{*}} \nabla \bar{v}_{*},
\end{equation}
where we can find that the main difference between $\nabla \bar{d}_{*}$ and $\nabla \bar{v}_{*}$ is the direction and the scaling factor ${\sqrt{t}}/{\bar{v}_{*}}$.
Typically, for a standard GPIS, the magnitude of the gradients at the surface does not have to be fixed.
However, the Eikonal equation~\eqref{eikonal} requires that the magnitude of the gradient is normalised to 1.
Since the gradient of $\bar{d}_{*}$ and $\bar{v}_{*}$ have opposite directions, we simply normalise the gradient of $\bar{v}_{*}$, and invert the sign to obtain $\ced{\nabla}\bar{d}_{*}$. 

Based on~\eqref{eq:gradient}, the predictive variance $\mathbb{V}\left[d_{*}\right]$ can be directly recovered in terms of $\mathbb{V}\left[v_{*}\right]$ using a first-order approximation, 
\begin{equation}
    \edit{\mathbb{V}\left[d_{*}\right]=\left(\frac{-\sqrt{t}}{\bar{v}_{*}}\right) \mathbb{V}\left[v_{*}\right]\left(\frac{-\sqrt{t}}{\bar{v}_{*}}\right)^{\top}}. 
\end{equation}

\subsection{Choice of Covariance Function}\label{covariance section}

As presented in our previous work~\cite{wu2021faithful}, we derived a covariance function that satisfies the screened Poisson~\eqref{eq:lim_heat_eq} which is a linear PDE.
\ced{Based on work presented in~\cite{hartikainen2010kalman}, it is possible to reformulate linear stochastic PDEs into GP regression models.
The method relies on matching the power spectral density of the covariance kernel and the PDE's finite-dimension Markov process.
Accordingly, using the} so-called Whittle kernel~\cite{Whittle}, \ced{from the Mat\'{e}rn family}, \edit{allows to recover a solution for} the screened Poisson equation~\eqref{eq:lim_heat_eq} as a GP regression model, %For brevity, we only discuss the 2D case. In the 2D case, we may expand \eqref{poissonE} as: % yes, it is the same v(x)
%\begin{equation}\label{eq:heat_pde_2d}
%    \frac{\partial^{2} v\left(x_{1}, x_{2}\right)}{\partial x_{1}^{2}}+\frac{\partial^{2} v\left(x_{1}, x_{2}\right)}{\partial x_{2}^{2}}-\lambda^{2} v\left(x_{1}, x_{2}\right)=w\left(x_{1}, x_{2}\right)
%\end{equation}
%where $w(x, y)$ is the white noise and $\lambda = 1/\sqrt{t}$. We take the Fourier transform of~\eqref{eq:heat_pde_2d} to get the spectral density,
%\begin{equation}\label{eq:spectral_density}
%S\left(\omega_{1}, \omega_{2}\right)=\frac{1}{\left(\omega_{1}^{2}+\omega_{2}^{2}+\lambda^{2}\right)^{2}}\,.
%\end{equation}
%Based on the Wiener-Khinchin theorem~\cite{GPbook}, the covariance function is given by the inverse Fourier transform of the spectral density~\eqref{eq:spectral_density}:
\begin{equation}
    k\left(\mathbf{x}, \mathbf{x}^{\prime}\right)=\frac{\left|\mathbf{x}-\mathbf{x}^{\prime}\right|}{2 \lambda} K_{\nu}\left(\lambda\left|\mathbf{x}-\mathbf{x}^{\prime}\right|\right)\,,
\end{equation}
where $\lambda = 1/\sqrt{t}$ and $K_{\nu}$ is the modified Bessel function of the second kind of order $\nu=1$.
\ced{From the GP regression viewpoint, }\edit{$\lambda$ is the} \ced{inverse of the} \edit{length scale hyperparameter of the covariance function.
It controls the smoothness and interpolation ability of the model.}
In this paper, hyperparameter optimisation is not considered.
\edit{Nevertheless,} \ced{from \eqref{eq:regularised_eikonal} we show that when $t$ gets closer to zero, corresponding to a larger $\lambda$, the proposed model} \edit{will produce a more accurate EDF.}
\edit{However,} \ced{a $\lambda$ too large would lead to the loss of} \edit{the GP's interpolation ability and cause numerical issues~\cite{wu2021faithful}.
On the other hand, a small $\lambda$ value, \ced{corresponding to a large length scale}, will introduce inaccuracies to the distance field} \ced{through the loss of details in the infered distance field.}
\edit{As the datasets used in this paper are indoor scenarios, we} \ced{address this} \edit{trade-off between interpolation and distance accuracy by setting $\lambda$ as 20.} %\edit{As $\lambda$ is a parameter adjusted according to application scenarios, }

\ced{Unfortunately, the Whittle covariance kernel is not differentiable as Mat\'{e}rn kernel functions are $\lvert \nu - 1 \rvert$ times differentiable.
Therefore, the Whittle kernel} cannot be used for gradient inference.
Given that the standard Whittle covariance is a special case of the Mat\'{e}rn family of covariance functions \ced{(with $\nu = 1$), we proposed in~\cite{wu2021faithful} to use the Mat\'{e}rn $\nu=3/2$ kernel function as it is the closest Mat\'{e}rn functions that is at least} once differentiable.
The Mat\'{e}rn family of covariance functions are given by:
\begin{equation}
    k(\mathbf{x}, \mathbf{x}^{\prime})=\sigma^{2} \frac{2^{1-\nu}}{\Gamma(\nu)}\left(\frac{\sqrt{2 \nu}}{l} |\mathbf{x}-\mathbf{x}^{\prime}|\right)^{\nu} K_{\nu}\left(\frac{\sqrt{2 \nu}}{l} |\mathbf{x}-\mathbf{x}^{\prime}|\right)
\end{equation}
where $l$ the characteristic length scale and $\sigma^{2}$ the signal's variance are the hyperparameters controlling the generalisation information of GP models, and $\nu$ manages the degree of smoothness.
%Sample functions from Mat\'{e}rn family are $|\nu-1|$ times differentiable. Thus, $\nu$ manages the degree of smoothness.
%\todo[inline]{Cedric: I might generate a small figure on Saturday to illustrate that and start the sentence with "As illustrated in Fig, we ..."}
We empirically observed in~\cite{wu2021faithful} that the Mat\'{e}rn $3/2$ function with $l={\sqrt{2 \nu}}/{\lambda}$ is a good approximation of Whittle covariance.% with the advantage that it is once differentiable. %Given that the modified Mat\'{e}rn $3/2$ allows \edit{gradient} prediction and Whittle does not, and the approximation is acceptable, we propose to use this Mat\'{e}rn version as a trade-off for extra information. 

Exploiting the accurate EDF with gradients in the space, not only around the boundary but also further away, establishes the mathematical foundation to solve the pose estimation problem.
Furthermore, faithfully modelling the EDF with gradients is sufficient for a trajectory optimisation-based planner, which requires the distance and gradients to the nearest obstacles to explore complex surroundings. Moreover, the GPIS provides continuous and probabilistic representation for mapping and surface reconstruction.
Our Log-GPIS unlocks the potential to provide a unified solution for sensor odometry, surface reconstruction, and safe navigation.

%Typically, consider we are given sparse measurements on surface for each frame, $\mathcal{X} = \{ \mathbf{x}_{i} \} \subset \partial S$, $i = 1 \ldots N$. 
%The aim of our representation is to approximate smooth and accurate $d(\mathbf{x})$ given $\mathcal{X}$, thereby reconstructing $S$ and performing the motion estimation.

%\section{Problem Definition}
\begin{figure*}[t]
	\centering
	\resizebox{0.9\linewidth}{!}{
	\includegraphics[]{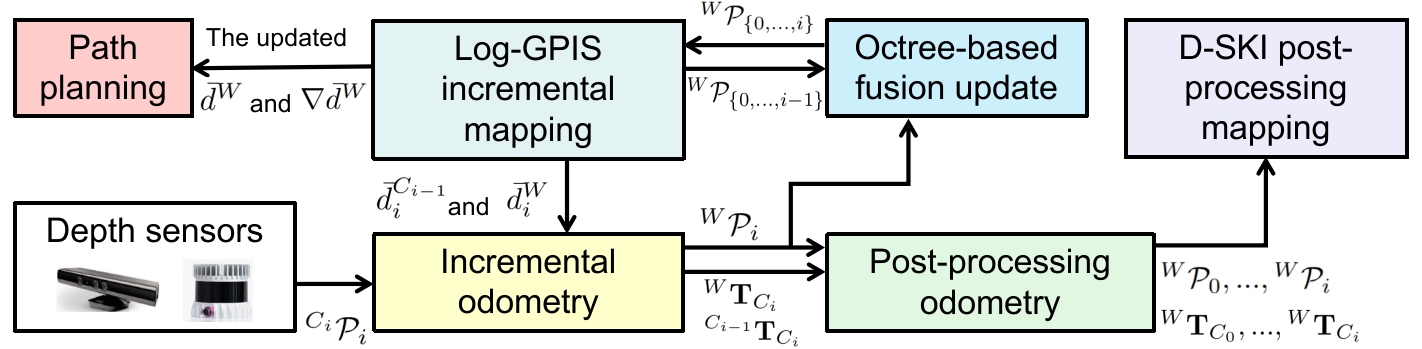}
	}
	\caption{\edit{Block Diagram of Log-GPIS-MOP framework.} \third{Depth sensors capture raw measurements as a sequence of point clouds ${ }^{{C}_{{i}}} \mathcal{P}_i$. Incremental odometry computes the frame-to-frame poses ${ }^{C_{i-1}} \mathbf{T}_{{C}_{{i}}}$ and frame-to-map poses ${ }^{W} \mathbf{T}_{{C}_{{i}}}$ based on the local Log-GPIS representation $\bar{d}_{i}^{C_{i-1}}$ and the global Log-GPIS representation $\bar{d}_{i}^W$. Applying the estimated frame-to-map pose, the current point cloud ${ }^{W} \mathcal{P}_i$ in the global frame then fuses with the existing points ${ }^{W} \mathcal{P}_{\{0,...,i-1\}}$ into ${ }^{W} \mathcal{P}_{\{0,...,i\}}$ to update the global Log-GPIS representation $\bar{d}_{i}^W$. The full updated representation $\bar{d}^W$ with gradients $\nabla \bar{d}^W$ is used for optimisation-based path planning. The post-processing odometry optimises a batch of poses and then the post-processing mapping uses a sequence of point clouds and optimised poses as input to create the map.}}
\vspace{-2ex}
	\label{fig:overview}
\end{figure*}

\section{Log-GPIS-MOP Overview}\label{method overview}

% problem formulation with Log-GPIS
Let us consider the Log-GPIS as a unified representation for simultaneous mapping, odometry and planning (MOP) with depth sensors. Log-GPIS-MOP aims \edit{to} incrementally estimate the EDF, while at the same time localising the sensor, and with this information plan its path to avoid colliding into the surfaces implicitly described by the EDF. \edit{Fig.~\ref{fig:overview} shows an overview of the full approach, with data flows and notations later explained in the Sec.~\ref{odometry section},~\ref{mapping section}, and~\ref{planning section}.} Given the input data from any type of depth sensor, e.g. LiDAR and depth cameras, the pose estimation problem for \edit{incremental} odometry is formulated as iterative alignment where the Log-GPIS \edit{incremental map} is used to minimise the distance from a newly arrived point cloud to the \edit{existing} surface following the gradients. The \edit{incremental mapping} consists in fusing the current point cloud into the existing Log-GPIS that probabilistically represents the EDF and its gradient. \edit{After a post-processing optimisation for a batch of poses, a post-processing mapping is applied based on the structured kernel interpolation framework with derivatives (D-SKI)~\cite{paper:D-SKI}.} Finally, a path planning approach uses the Log-GPIS \edit{incremental mapping} to avoid the surfaces of the mapped environment using the EDF and gradient information. Note that Log-GPIS-MOP is based on the assumption that the environment has sufficient curvature to unequivocally recover the pose of the sensor at any given sensor frame.

\section{Odometry}\label{odometry section}

%technically introduce the notation throuht the setup
Let us consider a depth sensor moving in a static environment. The sensor captures a sequence of organised or unorganised point clouds as measurements of the environment. Let us assume that the measurement noise is independent and Gaussian distributed. The sensor frame at time $t_i$ is denoted as $\mathfrak{F}_{{C}_{{i}}}$, where $i=(0,...,L)$. The world reference $\mathfrak{F}_{W}$ is given by the sensor frame $\mathfrak{F}_{{C}_{{0}}}$ at initial the timestamp. The pose of $\mathfrak{F}_{{C}_{{i}}}$ in world reference $\mathfrak{F}_{W}$ is given by the rotation matrix ${ }^{W} \mathbf{R}_{{C}_{{i}}} \in \operatorname{SO}(3)$ and the translation vector ${ }^{W} \mathbf{t}_{{C}_{{i}}}$, combined into the $4 \times 4$ homogeneous transformation matrix $\in \operatorname{SE}(3)$ as,
\begin{equation}
{ }^{W} \mathbf{T}_{{C}_{i}}=\left[\begin{array}{cc}
{ }^{W} \mathbf{R}_{{C}_{i}} & { }^{W} \mathbf{p}_{{C}_{i}} \\
\mathbf{0}^{\top} & 1
\end{array}\right]\,.
\end{equation}
%\begin{equation}
%{ }^{B} \mathbf{T}_{A}=\left[\begin{array}{cc}
%{ }^{B} \mathbf{R}_{A} & { }^{B} %\mathbf{p}_{A} \\
%\mathbf{0}^{\top} & 1
%\end{array}\right];
%\end{equation}

%\begin{equation}
%{{ }^{B} \mathbf{T}_{A}}^{-1}=\left[\begin{array}{cc}
%{{{ }^{B} \mathbf{R}_{A}}}^{\top} & -{{{ }^{B} \mathbf{R}_{A}}}^{\top} {{ }^{B} \mathbf{p}_{A}} \\
%\mathbf{0}^{\top} & 1
%\end{array}\right].
%\end{equation}

We denote the point cloud measurement of the sensor frame $\mathfrak{F}_{{C}_{{i}}}$ at time $t_i$ as ${ }^{{C}_{{i}}} \mathcal{P}_i$ and ${ }^{{C}_{{i}}}\mathbf{p}_{i,j} \in \mathbb{R}^{3}$ the $j$-th point of ${ }^{{C}_{{i}}} \mathcal{P}_i$ with $j = 0,...,J_{{i}}$. This 3D point cloud ${ }^{{C}_{{i}}} \mathcal{P}_i$ can be projected from $\mathfrak{F}_{{C}_{{i}}}$ to $\mathfrak{F}_{W}$ using ${ }^{W} \mathbf{T}_{{C}_{{i}}}$,
\begin{equation}
    \left[\begin{array}{c}
{ }^{W} \mathcal{P}_i \\
\mathbf{1}
\end{array}\right]={ }^{W} \mathbf{T}_{{C}_{{i}}}\left[\begin{array}{c}
{ }^{{C}_{{i}}} \mathcal{P}_i \\
\mathbf{1}
\end{array}\right]\,.
\end{equation}
Also, let us express ${ }^{W} \mathbf{T}_{{C}_{{i}}}$ through the concatenation of ${ }^{W} \mathbf{T}_{{C}_{{i-1}}}$ and ${ }^{{C}_{{i-1}}} \mathbf{T}_{{C}_{{i}}}$ as:
% \begin{equation}
%     T_{W C^{j}}=T_{W C^{i}} T_{C^{i} C^{j}},
% \end{equation}
\begin{equation}
    { }^{W} \mathbf{T}_{{C}_{{i}}}={ }^{W} \mathbf{T}_{{C}_{{i-1}}} { }^{{C}_{{i-1}}} \mathbf{T}_{{C}_{{i}}},
\end{equation}
where ${ }^{W} \mathbf{T}_{{C}_{{i-1}}}$ is given by the odometry computation at time $t_{i-1}$. We aim to find the ${ }^{W} \mathbf{T}_{{C}_{{i}}}$ such that the newly arrived ${ }^{{C}_{{i}}} \mathcal{P}_i$ can be projected to the world frame.

We present first the incremental formulation that leverages directly the Log-GPIS. This incremental odometry formulation is the one used by the incremental mapping and planning approaches all tightly integrated through the Log-GPIS. Furthermore, we also present the batch optimisation odometry formulation that can be used as a post-processing step to refine the trajectory estimation. Note that the batch formulation uses the sequential poses only and not directly the Log-GPIS.

% The transformation between coordinate frames $C^i$ and $C^j$ is denoted as $T_{C^{i} C^{j}} \in \operatorname{SE}(3)$. $T_{C^{i} C^{j}}$ corresponds points in frame $Cj$ to frame $W$ using:
% \begin{figure}[htb]
% 	\centering
% 	\resizebox{1\linewidth}{!}{
% 	\includegraphics[]{figures/factorgraph.png}
% 	}
% 	\caption{Factor graph illustration of the Log-GPIS odometry. Given a global representation merged from all previous frames, the distance residual of current sensor frame $C$ will be queried to compute the estimated rotation and translation. After poses obtained, update the global map $W$ through Bayesian fusion. }
% \vspace{-2ex}
% 	\label{factor}
% \end{figure}

%A collection of odometry-estimated poses is passed to the back-end directly as constraints. To avoid long time drifts and accumulated errors, a loop closure is detected from extra assistant to add additional constraint to the pose graph optimisation, which will correct and refine all the alignments by minimising the total non-linear least squares error. 
% Our sensor tracking method is based on Log-GPIS representation, so that we name it Log-GPIS odometry.

\subsection{Incremental Formulation}
In this work, we propose an iterative approach to estimate the odometry based on the Log-GPIS representation. \edit{Our odometry estimation approach follows an iterative optimisation to estimate the frame transformations. This is performed in a similar way to the Iterative Closest Point (ICP)\cite{ICP} but without looking for the closest points.}%Our odometry estimation approach follows the iterative and frame transformation estimation aspects behind Iterative Closest Point (ICP)\cite{ICP} but without the \edit{closest point (CP) searching}. ICP algorithm aligns two point clouds using, for instance, point-to-point or point-to-plane distance metrics. For each point in the source point cloud, ICP searches for the closest point in the target point cloud, which are tentative correspondences. Then, the transformation that aligns both point clouds is computed by iteratively minimising the total distance error of all points correspondences. \edit{As a crucial part of a standard ICP, searching for the correct and closest correspondence affects the ICP performance~\cite{kinectfusion,icp2}.}

Our work proposes instead, a direct query of the distance to the surface based on Log-GPIS to find the alignment efficiently and accurately without the need for the closest points search. Given a fused global representation with the information of all previous frames, the minimum distance to the surface can be obtained directly from the Log-GPIS, which saves us from doing the correspondences search. With the assumption that the difference between two consecutive frames is relatively small, the transformation is computed via a non-linear least-square optimisation involving the distances queried from the Log-GPIS effectively corresponding to point-to-plane constraints. After the transformation of the current frame is applied, the current point cloud is fused to the global representation using a fusion update method, which will be discussed in Sec.~\ref{fusion}.

%following the query gradients similar to a point to plane method
% \subsubsection{Frame to frame odometry}

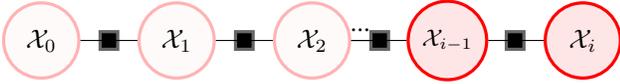
\begin{figure}[htb]
\centering
\begin{tikzpicture}[shorten >=0pt,node distance=0.9cm,on grid,
state/.style={circle, draw=red!100, fill=red!10, very thick, minimum size=10mm},
state1/.style={rectangle, draw=black!60, fill=black!100, very thick, minimum size=1mm},
state2/.style={circle, draw=red!30, fill=red!2, very thick, minimum size=10mm},
]
\node[state2] (C_0) {$\mathcal{X}_0$};
\node[state1](C_1) [right=of C_0] {$ $};
\node[state2] (C_2) [right=of C_1] {$\mathcal{X}_1$};
\node[state1](C_3) [right=of C_2] {$ $};
\node[state2] (C_4) [right=of C_3] {$\mathcal{X}_2$};
\node[state1](C_5) [right=of C_4] {$ $};
\node[state] (C_6) [right=of C_5] {\small$\mathcal{X}_{i-1}$};
\node[state1](C_7) [right=of C_6] {$ $};
\node[state] (C_8) [right=of C_7] {$\mathcal{X}_i$};
% \node[state1] (a) at (2.7,0.5) {$ $};
% \node[state1] (b) at (4.5,0.5) {$ $};
% \node[state1] (c) at (6.3,0.5) {$ $};
\path[-]  (C_0) edge              node [above] {} (C_2)
                %edge [bend left]  node [below] {} (C_2)
                %edge [bend left]  node [above] {} (C_4)
                %edge [bend left]  node [below] {} (C_6)
                %edge [bend left]  node [above] {...} (C_8)
          (C_2) edge              node [above] {} (C_4)
          (C_4) edge              node [above] {} (C_6)
          %(C_4) edge              node [above] {...} (C_5);
          (C_6) edge              node [above] {} (C_8)
          (C_4) edge              node [above] {...} (C_5);
\end{tikzpicture}
\caption{Factor graph representation of the frame-to-frame odometry. \third{$\mathcal{X}_i=\{{ }^{W} \mathbf{T}_{{C}_{{1}}},...,{ }^{W} \mathbf{T}_{{C}_{{i}}}\}$ with $i=(0,...,L)$.} Low opacity nodes are used for those nodes that are not used for current alignment. Black factors are the local Log-GPIS distance constraints.} 
\label{odometry f2f}
\end{figure}

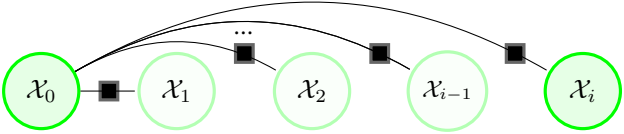
\begin{figure}[htb]
\centering
\begin{tikzpicture}[shorten >=1pt,node distance=0.9cm,on grid,
state/.style={circle, draw=green!100, fill=green!10, very thick, minimum size=10mm},
state1/.style={rectangle, draw=black!60, fill=black!100, very thick, minimum size=1mm},
state2/.style={circle, draw=green!30, fill=green!2, very thick, minimum size=10mm},
]
\node[state] (C_0) {$\mathcal{X}_0$};
\node[state1](C_1) [right=of C_0] {$ $};
\node[state2] (C_2) [right=of C_1] {$\mathcal{X}_1$};
%\node[state1](C_3) [right=of C_2] {$ $};
\node[state2] (C_4) [right=of C_3] {$\mathcal{X}_2$};
%\node[state1](C_5) [right=of C_4] {$ $};
\node[state2] (C_6) [right=of C_5] {\small$\mathcal{X}_{i-1}$};
%\node[state1](C_7) [right=of C_6] {$ $};
\node[state] (C_8) [right=of C_7] {$\mathcal{X}_{i}$};
\node[state1] (a) at (2.7,0.5) {$ $};
\node[state1] (b) at (4.5,0.5) {$ $};
\node[state1] (c) at (6.3,0.5) {$ $};
\path[-]  (C_0) edge              node [above] {} (C_2)
                %edge [bend left]  node [below] {} (C_2)
                edge [bend left]  node [above] {} (C_4)
                edge [bend left]  node [below] {} (C_6)
                edge [bend left]  node [below] {...} (C_6)
                edge [bend left]  node [above] {} (C_8);
          %(C_2) edge              node [above] {} (C_4)
          %(C_4) edge              node [above] {} (C_6)
          %(C_6) edge              node [above] {} (C_8)
          %(C_6) edge              node [above] {...} (C_7);
%\path[]  (C_4) edge              node [above] {...} (C_6);
\end{tikzpicture}
\caption{\third{Factor graph for the frame-to-map odometry. Black factors represent the global Log-GPIS distance constraints.}} \label{odometry f2m}
\end{figure}

More formally, let us consider a local Log-GPIS representation \edit{$\bar{d}_{i}^{C_{i-1}}$} obtained using the previous frame only and a global Log-GPIS representation $\bar{d}_{i}^W$, which is the result of fusing all the previous frames into the \edit{world} frame. Let us assume the initial ${ }^{W} \mathbf{T}_{{C}_{{0}}}$ is identity and define \third{$\mathcal{X}_i=\{{ }^{W} \mathbf{T}_{{C}_{{1}}},...,{ }^{W} \mathbf{T}_{{C}_{{i}}}\}$} with $i=(0,...,L)$ as the state to be estimated. The odometry is then formulated incrementally to estimate ${ }^{W} \mathbf{T}_{{C}_{{i}}}$ through a frame-to-frame alignment Fig.~\ref{odometry f2f}, with the possibility to compute the frame-to-map alignment as shown in~Fig.~\ref{odometry f2m}.
% $\{{ }^{W} \mathcal{P}_0,...,{ }^{W} \mathcal{P}_i\}$

The iterative alignment in both cases, frame-to-map and frame-to-frame, is performed by minimising a cost function \third{$e^i_{\mathrm{dis}}$},
% \begin{equation}
% \hat{{ }^{W} \mathbf{T}_{{C}_{{i}}}} = \underset{{ }^{W} \mathbf{T}_{{C}_{{i}}}}{\arg \min }  \left\|e_{\mathrm{dis}}\left({ }^{W} \mathbf{T}_{{C}_{{i}}}\right)\right\|^{2},
% \end{equation}
% \begin{equation} \label{eq:edis}
% \check{\mathcal{X}_i} = \underset{\mathcal{X}_i}{\arg \min }  \left\|e^i_{\mathrm{dis}}\right\|_{\edit{\sigma^i_{dis}}}^{2},
% \end{equation}
\begin{equation} \label{eq:edis}
\check{\mathcal{X}_i} = \underset{\mathcal{X}_i}{\arg \min } \; \edit{e^i_{\mathrm{dis}}},
\end{equation}
% \begin{equation}
%     C(\mathcal{X}) = \sum \lVert e_{\mathrm{dis}}\left(\mathcal{X}\right)\lVert^2_{\boldsymbol{\Omega}_{dis}},
% \end{equation}
where $e^i_{\mathrm{dis}}$ \edit{is a scalar and} corresponds to the sum of the squared distances from \edit{each points in} the current point cloud ${ }^{{C}_{{i}}} \mathcal{P}_i$ to \edit{the surface representation Log-GPIS. Frame-to-frame case uses the local surface representation and frame-to-map case uses the global surface representation.} $\check{\mathcal{X}}_i$ is the sequential estimation of the current pose. %More importantly, \edit{the Log-GPIS representation} is directly used to compute \edit{the total distance} from \edit{all the points} in current point cloud ${ }^{{C}_{{i}}} \mathcal{P}_i$ by querying the distance \edit{mean value and variance}.

\paragraph{Frame-to-frame} 
\edit{This alignment is computed through the current frame $i$ querying the distance from the representation built from the previous sensor frame $C_{i-1}$ only (the local Log-GPIS $\bar{d}_{i}^{C_{i-1}}$).}

% tikz figures

Particularising \eqref{eq:edis} for this case (Fig.~\ref{odometry f2f}), let us write $e_{\mathrm{dis}}^{i}$ as: 
\begin{equation}
\begin{aligned}
    e_{\mathrm{dis}}^{i}\left({ }^{C_{i-1}} \mathbf{T}_{{C}_{{i}}}\right)=
    %{ }^{W} \mathbf{T}_{{C}_{{i-1}}}
    \sum_{j=0}^{J_{i}} \edit{\left\|\bar{d}_{i}^{C_{i-1}}\left({ }^{{C}_{{i-1}}} \mathbf{T}_{{C}_{{i}}}{ }^{C_{i}}  \mathbf{p}_{i,j}\right)\right\|_{\edit{\sigma^i_{dis}}}^{2}},
\end{aligned}
\end{equation}
%where the ${}^{W}\mathbf{T}_{{C}_{{i-1}}}$ is given by accumulating the relative poses from $0$ to $i-1$.
%
% where $r_{i,j}^{i-1}$ is the residual of each querying point given local representation and $r_{i,j}^{W}$ is the residual with the global representation.
%
% \begin{equation}\label{residual_frame2frame}
% \begin{aligned}
% r_{i,j}^{i-1}\left({ }^{{C}_{{i}}} \mathbf{x}_{j}, { }^{C_{i-1}} \mathbf{T}_{{C}_{{i}}}\right) %&=\psi_{{C}^{j}}\left(\mathbf{x}_{{C}^{j}}^{m}\right)-\psi_{{C}^{i}}\left(T_{C^{i} C^{j}} \mathbf{x}_{{C}^{j}}^{m}\right) \\
% % &=\bar{d}_{i-1}\left({ }^{{C}_{{i-1}}}  \mathbf{T}_{{C}_{{i}}}{ }^{{C}_{{i}}} \mathbf{x}_{j}\right) \\
% &=\bar{d}_{i-1}\left({ }^{{C}_{{i-1}}} \mathbf{T}_{{C}_{{i}}}{ }^{C_{i}}  \mathbf{x}_{j}\right)=\bar{d}_{i-1}\left({ }^{C_{i-1}}  \mathbf{x}_{j}\right),
% \end{aligned}
% \end{equation}
\edit{where $\sigma^i_{dis}$ is the distance variance from the Log-GPIS inference.} Given the local Log-GPIS for an arbitrary testing point ${ }^{{C}_{{i-1}}} \mathbf{p}_{i,j}$, the predictive mean \edit{$\bar{v}_{i}^{C_{i-1}}$} with derivatives are obtained from \eqref{eq:gp_inference_gradients}. Also, as per \eqref{3}, $\bar{d}_{i}^{C_{i-1}}$ can be computed by applying the logarithm transformation to \edit{$\bar{v}_{i}^{C_{i-1}}$},
\begin{equation}
\begin{aligned}
\begin{bmatrix}
\edit{\bar{v}_{i}^{C_{i-1}}} \\
\edit{\nabla \bar{v}_{i}^{C_{i-1}}}
\end{bmatrix} &=
 (\tilde{\mathbf{k}}\edit{^{C_{i-1}})}^{\top}(\tilde{K}\edit{^{C_{i-1}}}+\sigma^2 I)^{-1}\begin{bmatrix}
\mathbf{y}\edit{^{C_{i-1}}} \\
\nabla \mathbf{y}\edit{^{C_{i-1}}}
\end{bmatrix}
\end{aligned}
\end{equation}

\begin{equation}
    \edit{\bar{d}_{i}^{C_{i-1}}}=-\sqrt{t} \ln \edit{\bar{v}_{i}^{C_{i-1}}}.
\end{equation}
\edit{$\tilde{K}^{C_{i-1}}$} is the covariance matrix of the input points \edit{${ }^{{C}_{{i-1}}} \mathcal{P}_{i-1}$} and \edit{$\tilde{\mathbf{k}}^{C_{i-1}}$} represents the covariance vector between the input points and the testing point \edit{${ }^{{C}_{{i-1}}} \mathbf{p}_{i,j}$}. \edit{$\mathbf{y}^{C_{i-1}}$} are the measurements of ${ }^{{C}_{{i-1}}} \mathcal{P}_{i-1}$ of frame $\mathfrak{F}_{{C}_{{i-1}}}$ at $t_{i-1}$ and \edit{$\nabla \mathbf{y}^{C_{i-1}}$ are the noisy surface normals computed from the pointcloud (normal computation is further explained in Sec.~\ref{fusion}).} %, \emph{i.e.} ${ }^{{C}_{{i-1}}} \mathcal{P}_{i-1}$. 

As in any frame-to-frame odometry, it can easily drift. The other downside of this simple alignment is that the local log-GPIS has to be created at every step and discarded afterwards. This motivates the use of a more robust frame-to-map alignment, which uses the fused global Log-GPIS.

\paragraph{Frame-to-Map}

Let us consider a sequentially computed global Log-GPIS $\bar{d}_{i}^W$, which uses frame-to-map alignment to compute the poses as the sensor moves through the environment. \edit{The global Log-GPIS representation is the result of fusing all the previous frames into a single representation in the world coordinate frame.} Particularising $e_{\mathrm{dis}}^{i}$ for frame-to-map odometry (Fig.~\ref{odometry f2m}), we have:
% \begin{equation}\label{residual_frame2map}
% \begin{aligned}
% r_{i,j}^{W}\left({ }^{{C}_{{i}}} \mathbf{x}_{j}, { }^{W} \mathbf{T}_{{C}_{{i}}}\right) %&=\psi_{{C}^{j}}\left(\mathbf{x}_{{C}^{j}}^{m}\right)-\psi_{{C}^{i}}\left(T_{C^{i} C^{j}} \mathbf{x}_{{C}^{j}}^{m}\right) \\
% &=\bar{d}_{W}\left({ }^{W} \mathbf{T}_{{C}_{{i}}}{ }^{{C}_{{i}}} \mathbf{x}_{j}\right) =\bar{{d}_{W}}\left({ }^{W}  \mathbf{x}_{j}\right),
% \end{aligned}
% \end{equation}
\begin{equation}
\begin{aligned}
    e_{\mathrm{dis}}^{i}\left({ }^{W} \mathbf{T}_{{C}_{{i}}}\right)&=
    \sum_{j=0}^{J_{i}} \edit{\left\|\bar{d}_{i}^W\left({ }^{W} \mathbf{T}_{{C}_{{i}}}{ }^{{C}_{{i}}} \mathbf{p}_{i,j}\right)\right\|_{\sigma^i_{dis}}^{2}},
\end{aligned}
\end{equation}
where the global Log-GPIS $\bar{d}_{i}^W$ is computed from,

\begin{equation}
\begin{aligned}
\begin{bmatrix}
\bar{v}_{i}^W \\
\nabla \bar{v}_{i}^W
\end{bmatrix} &=
 (\tilde{\mathbf{k}}^{W})^{\top}(\tilde{K}^{W}+\sigma^2 I)^{-1}\begin{bmatrix}
\mathbf{y}^{W} \\
\nabla \mathbf{y}^{W}
\end{bmatrix}
\end{aligned}
\end{equation}
\begin{equation}
    \bar{d}_{i}^W=-\sqrt{t} \ln \bar{v}_{i}^W,
\end{equation}
with $\tilde{K}^{W}$ the covariance matrix of all the fused point clouds from $0$ to $i-1$ in \edit{world} reference frame ${ }^{W} \mathcal{P}_{\{0,...,i-1\}}$ and $\tilde{\mathbf{k}}^{W}$ \edit{the covariance vector between ${ }^{W} \mathcal{P}_{\{0,...,i-1\}}$ and the testing point \edit{${ }^{W} \mathbf{p}_{i,j}$}. $\mathbf{y}^{W}$ and $\nabla \mathbf{y}^{W}$ are the transformed implicit surface value and the normal computed from the pointcloud of ${ }^{W} \mathcal{P}_{\{0,...,i-1\}}$.}

%\paragraph{Jacobian for Frame to Map}
Further, to solve the optimisation of the alignment problem, the derivative of the distance with respect to ${ }^{W} \mathbf{T}_{{C}_{{i}}}$ is desired. Based on the chain rule, the Jacobian function is as follows,
\begin{equation}
    \begin{aligned}
    \frac{\partial\bar{d}_{i}^W\left({ }^{W} \mathbf{T}_{{C}_{{i}}}{ }^{{C}_{{i}}} \mathbf{p}_{i,j}\right)}{\partial \left({ }^{W} \mathbf{T}_{{C}_{{i}}}\right)} =\left.\frac{\partial \bar{d}_{i}^W\left({ }^{W} \mathbf{T}_{{C}_{{i}}}{ }^{{C}_{{i}}} \mathbf{p}_{i,j}\right)}{\partial \left({ }^{W} \mathbf{T}_{{C}_{{i}}}{ }^{{C}_{{i}}} \mathbf{p}_{i,j}\right)}\right. \frac{\partial\left({ }^{W} \mathbf{T}_{{C}_{{i}}}{ }^{{C}_{{i}}} \mathbf{p}_{i,j}\right)}{\partial \left({ }^{W} \mathbf{T}_{{C}_{{i}}}\right)}
\end{aligned}
\end{equation}
%\subsection{First term of Jacobian}

The first term is one of the valuable byproducts of the global Log-GPIS, which is the gradient of query point ${ }^{W} \mathbf{T}_{{C}_{{i}}}{ }^{{C}_{{i}}} \mathbf{p}_{i,j}$: 
\begin{equation}
    \left.\left.\frac{\partial \bar{d}_{i}^W\left({ }^{W} \mathbf{T}_{{C}_{{i}}}{ }^{{C}_{{i}}} \mathbf{p}_{i,j}\right)}{\partial \left({ }^{W} \mathbf{T}_{{C}_{{i}}}{ }^{{C}_{{i}}} \mathbf{p}_{i,j}\right)}\right.\right. = \nabla \bar{d}_{i}^W({ }^{W} \mathbf{T}_{{C}_{{i}}}{ }^{{C}_{{i}}} \mathbf{p}_{i,j}).
\end{equation}
The second term is the derivative of ${ }^{W} \mathbf{T}_{{C}_{{i}}}$ with respect to the point ${ }^{{C}_{{i}}} \mathbf{p}_{i,j}$. It is equivalent to
\begin{equation}
    \frac{\partial\left({ }^{W} \mathbf{T}_{{C}_{{i}}}{ }^{{C}_{{i}}} \mathbf{p}_{i,j}\right)}{\partial \left({ }^{W} \mathbf{T}_{{C}_{{i}}}\right)} = \left[\begin{array}{cc}
\boldsymbol{I} & -\left({ }^{W} \mathbf{T}_{{C}_{{i}}}{ }^{{C}_{{i}}} \mathbf{p}_{i,j}\right)^{\wedge} \\
\mathbf{0}^{T} & \mathbf{0}^{T}
\end{array}\right],
\end{equation}
where $\wedge$ is the skew-symmetric operator for the non-homogeneous 3D point $\mathbf{p}$,
\begin{equation}
     \mathbf{p}^{\wedge}=\left[\begin{array}{l}
{p}_{1} \\
{p}_{2} \\
{p}_{3}
\end{array}\right]^{\wedge}=\left[\begin{array}{ccc}
0 & -{p}_{3} & {p}_{2} \\
{p}_{3} & 0 & -{p}_{1} \\
-{p}_{2} & {p}_{1} & 0
\end{array}\right]\,.
\end{equation}

%Combining the two terms, the Jacobian function will be a $1 \times 6$ vector $\in \mathfrak{s o}(3)$. 
%With the last rows removed, we will have a 3 by 6 matrix for the second term. Moreover, the first term is a 1 by 3 vector. 
Using the querying distance and analytical Jacobian function, we implement our proposed odometry in Ceres \footnote{\url{https://ceres-solver.org}}. %The Ceres will iteratively finish the process and find the alignment when the solver meets the termination requirements.

\subsection{Post Process Batch Optimisation}

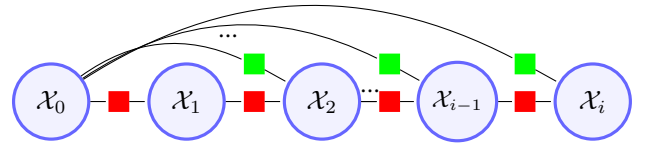
\begin{figure}[htb]
\centering
\begin{tikzpicture}[shorten >=1pt,node distance=0.9cm,on grid,
state/.style={circle, draw=blue!60, fill=blue!5, very thick, minimum size=10mm},
state1/.style={rectangle, draw=red!100, fill=red!100, very thick, minimum size=1mm},
state2/.style={rectangle, draw=green!100, fill=green!100, very thick, minimum size=1mm},
]
\node[state] (C_0) {$\mathcal{X}_0$};
\node[state1](C_1) [right=of C_0] {$ $};
\node[state] (C_2) [right=of C_1] {$\mathcal{X}_1$};
\node[state1](C_3) [right=of C_2] {$ $};
\node[state] (C_4) [right=of C_3] {$\mathcal{X}_2$};
\node[state1](C_5) [right=of C_4] {$ $};
\node[state] (C_6) [right=of C_5] {\small$\mathcal{X}_{i-1}$};
\node[state1](C_7) [right=of C_6] {$ $};
\node[state] (C_8) [right=of C_7] {$\mathcal{X}_{i}$};
\node[state2] (a) at (2.7,0.5) {$ $};
\node[state2] (b) at (4.5,0.5) {$ $};
\node[state2] (c) at (6.3,0.5) {$ $};
\path[-]  (C_0) edge              node [above] {} (C_1)
                edge [bend left]  node [above] {} (a)
                edge [bend left]  node [below] {...} (b)
                edge [bend left]  node [below] {} (c)
          (a)   edge              node [above] {} (C_4)
          (b)   edge              node [above] {} (C_6)
          (c)   edge              node [above] {} (C_8)
          (C_2) edge              node [above] {} (C_3)
          (C_3) edge              node [above] {} (C_4)
          (C_4) edge              node [above] {} (C_5)
          (C_5) edge              node [above] {} (C_6)
          (C_6) edge              node [above] {} (C_7)
          (C_7) edge              node [above] {} (C_8)
          (C_4) edge              node [above] {...} (C_5)
          (C_1) edge              node [above] {} (C_2);
\end{tikzpicture}
\caption{\third{Factor graph representation of the pose SLAM problem. Green factors represent the frame-to-map constraints and red factors are the frame-to-frame constraints.}} \label{odometry f2f f2m}
\end{figure}

Given the incremental pose estimation with frame-to-frame and frame-to-map transformations, it is possible to optimise \edit{a batch of poses through a pose graph optimisation (PGO)~\cite{Cadena16tro-SLAMfuture,grisetti2010tutorial,bai2021sparse}. \edit{\third{Note that the batch optimisation presented is an offline process to refine the map, and it is not used during incremental mapping and planning.}} As shown in Fig.~\ref{odometry f2f f2m}, blue circles are the vertices representing poses. Green and red factors are the constrained edges, which represent the relations between poses. The edges are computed from the frame-to-map and frame-to-frame odometry respectively. We employ the standard PGO formulation as a Maximum Likelihood Estimation (MLE)~\cite{Cadena16tro-SLAMfuture,grisetti2010tutorial,bai2021sparse}}:
% \begin{equation}
%     \hat{\mathcal{X}}=\underset{\mathcal{X}}{\operatorname{argmin}}-\log (p(\mathcal{X} \mid \check{\mathcal{X}}))= \underset{\mathcal{X}}{\arg \min }  \left\|e_{\mathrm{pgo}}\right\|^{2}_{\Sigma_{\mathrm{pgo}}},%=\underset{\mathcal{X}}{\operatorname{argmin}} C(\mathcal{X}),
% \end{equation}
\begin{equation}
    \hat{\mathcal{X}}=\underset{\mathcal{X}}{\operatorname{argmin}}-\log (p(\mathcal{X} \mid \check{\mathcal{X}}))= \underset{\mathcal{X}}{\arg \min }  \; \edit{e_{pgo}},%=\underset{\mathcal{X}}{\operatorname{argmin}} C(\mathcal{X}),
\end{equation}
where $\hat{\mathcal{X}}$ is the state of poses and, $e_{pgo}$ is the error of the poses. Here, the PGO takes the frame-to-map constraints (in green measurements) and frame-to-frame constraints (in red measurements) to \edit{formulate the $e_{pgo}$~\cite{Cadena16tro-SLAMfuture,bai2021sparse}:}
\begin{equation}
    {e}_{\text {pgo}}\left({ }^{W} \mathbf{T}_{{C}_{{i}}}\right)= \sum_{i \in 0,.,L} \edit{\left\|\log \left({ }^{W} \check{\mathbf{T}}_{{C}_{{i}}}^{-1} { }^{W} \mathbf{T}_{{C}_{{i}}}\right)\right\|^{2}_{\Sigma^{i}_{gpo}}}\,,
\end{equation}
\edit{where ${ }^{W} \check{\mathbf{T}}_{{C}_{{i}}}^{-1}$ is the noisy estimation of ${ }^{W} \mathbf{T}_{{C}_{{i}}}$. $\Sigma^i_{pgo}$ is the covariance of poses.}

\section{Mapping}\label{mapping section}
\edit{We present two different pipelines, one online and one offline. The incremental mapping aims to sequentially fuse new incoming point clouds into the global Log-GPIS representation for odometry and planning. The post-processing mapping uses the output of the \edit{post-processing} odometry and the raw point clouds to recover the full \edit{reconstruction} for visualisation.}

\subsection{Incremental Mapping}\label{fusion}
After the pose is estimated and applied as described in Sec.~\ref{odometry section}, the current frame needs to be fused/appended with the global Log-GPIS representation \edit{for incremental mapping}. To do so, we adopt the approach proposed in~\cite{Bhoram} for GPIS-SDF. This approach uses an Octree-based Gaussian Process regressor and Bayesian fusion to merge old measurements and new observations. 

We briefly summarise the process here, which has been generalised for organised and unorganised point clouds. \edit{To initially remove outliers, we follow the neighbourhood filtering technique in~\cite{points_denoise} reducing excessive noise without penalising important details. Moreover, we set a minimum and maximum range for the sensor data. Points beyond the maximum range and closer than the minimum range are considered invalid.} Let us then consider the current point cloud ${ }^{W} \mathcal{P}_i$ and all fused global points ${ }^{W} \mathcal{P}_{\{0,...,i-1\}}$. \edit{The objective of incremental mapping is to have the new fused global point cloud ${ }^{W} \mathcal{P}_{\{0,...,i\}}$}. The full point cloud ${ }^{W} \mathcal{P}_{\{0,...,i-1\}}$ is stored in an Octree-based structure to reduce the computational complexity, which divides the complete data into many overlapping clusters. \edit{The clusters within the same region as ${ }^{W} \mathcal{P}_i$ are set as active clusters and needs to be fused with ${ }^{W} \mathcal{P}_i$.} In order to fuse ${ }^{W} \mathcal{P}_i$ with active clusters in ${ }^{W} \mathcal{P}_{\{0,...,i-1\}}$, let us model a GP regressor \edit{using ${ }^{W} \mathcal{P}_i$} in terms of the bearing angles $\alpha$ and $\varepsilon$ of each point $\third{{ }^{W}\mathbf{p}} \in { }^{W} \mathcal{P}_i$ and its inverse range for either unorganised or organised point clouds as,
% \begin{equation}
%     \rho^{-1}=\gamma(\alpha,\varepsilon).
% \end{equation}
\begin{equation}\label{regressor}
    \rho^{-1} \sim \mathcal{GP}\{{0}, k((\alpha,\varepsilon), (\alpha',\varepsilon'))\}.
\end{equation}
The Ornsten-Uhlenbeck covariance\edit{~\cite{GPbook}} is used in this case as it captures more details without strong smoothing. Note that for organised point clouds the conversion from pixel coordinates to bearing angles given the camera calibration is straightforward. Then GP regressor can infer smoothly and probabilistically the reciprocal range value at any given point with values $\alpha$ and $\varepsilon$. \third{Each active point ${ }^{W} \mathbf{p'}$ in the Octree of the existing representation ${ }^{W} \mathcal{P}_{\{0,...,i-1\}}$ needs to be fused with ${ }^{W} \mathcal{P}_{i}$ using the above GP regressor.} 

The regressor accepts the bearings of \third{${ }^{W} \mathbf{p'}$} to infer the range value $\bar{\rho}^{-1}$ as in \eqref{regressor}. Then, we test the occupancy function $\Phi$ \third{for ${ }^{W} \mathbf{p'}$} below,
\begin{equation}\label{occupancy}
    % \Phi = 1-\frac{2}{1+\exp \left\{-a\left[\delta(\mathbf{x}_{W}^{m})^{-1}-r\left(\theta(\mathbf{x}_{W}^{m})\right)\right]\right\}}
    \Phi = \frac{2}{1+\exp \left(-\eta\left(\rho'^{-1}-\bar{\rho}^{-1}\right)\right)},
\end{equation}
where $\rho'$ is the given range of the point and $\bar{\rho}$ is the range inference. $\eta$ is a slope parameter. The point is free with $\rho'>\bar{\rho}$, and occupied with $\rho'<\bar{\rho}$.

If the occupancy value is larger than a threshold, then the target point moving procedure is performed. With a negative occupancy value beyond the surface, it moves one step along the direction of the surface normal. On the contrary, with a positive value, it moves in the opposite direction of normal. Iteratively, using the regressor, the algorithm ends up with a new point \third{${{ }^{W} {\mathbf{p''}}}$} very close to the surface. \edit{The normal of \third{${{ }^{W} {\mathbf{p''}}}$} is computed using $(\Phi\left(\third{{{ }^{W} {\mathbf{p''}}}}+\delta \boldsymbol{e}_v\right)-\Phi\left(\third{{{ }^{W} {\mathbf{p''}}}}-\delta \boldsymbol{e}_v\right))/{2 \delta}$, where $\delta$ is a small positive value and $\boldsymbol{e}_v$ is unit vector along each normal axis.}

Given \third{${ }^{W} \mathbf{p'}$} with variance \third{$\mathbf{\sigma'}$} \edit{(in the existing global Log-GPIS)} and the inferred point \third{${{ }^{W} {\mathbf{p''}}}$} with variance \third{$\mathbf{\sigma''}$} \edit{(in the current frame)} both in the overlapping part, the Bayesian fusion directly updates the surface point \third{${ }^{W} \mathbf{p''}$} with variance \third{$\mathbf{\sigma''}$} \edit{that implicitly will update each points in active clusters for the global Log-GPIS representation} using:
\third{
\begin{equation}
    \begin{aligned}
{ }^{W} {\mathbf{p''}} &\gets\frac{\mathbf{\sigma''} { }^{W} \mathbf{p'}+\mathbf{\sigma'} {{ }^{W} {\mathbf{p''}}}}{\mathbf{\sigma''}+\mathbf{\sigma'}} \\
% \mathbf{\xi} &=\frac{\mathbf{\xi}_{W}^{m}+\mathbf{\xi}_{{C}^{j}}^{m}}{\mathbf{\xi}_{W}^{m}+\mathbf{\xi}_{{C}^{j}}^{m}} 
(\mathbf{\sigma''})^{-1} &\gets(\mathbf{\sigma''})^{-1}+(\mathbf{\sigma'})^{-1}\,.
\end{aligned}
\end{equation}}
% In addition, with \eqref{occupancy}, the gradient
% \begin{equation}
%     \frac{\partial \tilde{x}_{k}}{\partial v} \approx \frac{\operatorname{occ}\left(\tilde{x}_{k}+\delta \boldsymbol{e}_{v}\right)-\operatorname{occ}\left(\tilde{x}_{k}-\delta \boldsymbol{e}_{v}\right)}{2 \delta}
% \end{equation}

\subsection{Post Processing Mapping}
Given the final estimated transformations ${ }^{W} \mathbf{T}_{{C}_{{i}}}$ for the sequence of point clouds ${ }^{{C}_{{i}}} \mathcal{P}_i$, which can be naively merged into one point cloud in the world reference frame, we aim to build a globally consistent continuous and probabilistic reconstruction. However, a well-known disadvantage of general GPs lies in the memory allocation and the computational complexity to invert the covariance matrix of size $J$ the number of training points. Moreover, as pointed out above, we model the joint GPIS with gradients, which improves the accuracy of the predicted surfaces. However, considering the gradients as the input increases the size of the covariance matrix by a factor of four. This significantly worsens the computational cost of the plain GPIS. %Despite this being advantageous as there is no need to generate artificial points to direct inside and outside surfaces, the computational complexity gets worse when considering the gradient inferences. 

To reduce the general computational complexity, and thereby improve scalability, in our previous work~\cite{LanRAL20}, \edit{we adopted the idea of the Structured Kernel Interpolation algorithm (SKI)~\cite{paper:KISS-GP} to approximate the exact kernel matrix through interpolation weights and inducing points, and extend to D-SKI when considering the derivatives with SKI~\cite{paper:D-SKI,paper:Kun}.} The input dataset is reprojected onto a grid generated by inducing points, which significantly reduces the requirement to compute GPIS. The interpolated \edit{covariance} function of $k\left(\mathbf{x}, \mathbf{x}'\right)$ is formulated as,
% \begin{equation} \label{SKI}
%  k\left(\mathbf{x}, \mathbf{x}'\right) \approx W k(\mathbf{v},\mathbf{v}') W^{T},
% \end{equation}
\begin{equation} \label{SKI}
 K \approx V K(U,U) V^{T},
\end{equation}
where $U$ is a uniform grid of inducing points $m$. \edit{$V$ is a $\edit{J} \times m$ sparse matrix of weights used for interpolation, which intuitively represents the relative distances from input points to inducing points~\cite{paper:KISS-GP}. Formally, we use quintic~\cite{paper:Quintic} interpolation to obtain V.} When considering gradients and preserving the positive definiteness of the approximate kernel, D-SKI differentiates the combined covariance matrix. The \edit{approximate} covariance matrix \edit{with derivatives} of $\tilde{k}(\mathbf{x}, \mathbf{x}')$ is:
\begin{multline} \label{D-SKI}
 \tilde{K} \approx \left[\begin{array}{c}{V} \\ {\nabla V}\end{array}\right] K(U,U)\left[\begin{array}{c}{V} \\ {\nabla V}\end{array}\right]^{T}=\\\
\left[\begin{array}{cc}{V K(U,U) V^{T}} & {V K(U,U)(\nabla V)^{T}} \\ {(\nabla V) K(U,U) V^{T}} & {(\nabla V) K(U,U)(\nabla V)^{T}}\end{array}\right] \,,
\end{multline}
where $\partial V$ is the derivative of $V$.% with quintic interpolation~\cite{paper:Quintic}.

\subsection{Marching Cubes for Log-GPIS}
\edit{As we mentioned in Sec.~\ref{sec:derivation}, the surface is no longer at the zero crossing and the sign information is lost. Thus, to obtain the mesh for dense reconstruction, it requires a modified version of the marching cubes algorithm~\cite{marching}.} Given a set of query points for Log-GPIS to infer the implicit surface values, the \third{marching cubes} compute and extracts the surface by connecting points of a constant value (iso-value) within a volume of space:
\begin{equation}
    \mathbf{FV} = {f}_{iso}(\mathcal{M},v_{iso}),
\end{equation}
where $\mathcal{M}$ specifies a set of points for querying. $v_{iso}$ is the iso-value at which to compute the surface, specified as a scalar. $\mathbf{FV}$ contains the computed faces and vertices of the surface. Note that since the sign is lost, instead of going through exactly zero crossing, our marching cubes sets the iso-value as a constant value ($0.1$m for 2D cases and $0.001$m for 3D in our experiments). Then, the computed vertices in $\mathbf{FV}$ move towards the implicit surface and iteratively query the Log-GPIS until \edit{the local minimum is found} (iso-values are smaller than a threshold. We set $0.0001$m in our experiments).

\section{Planning}\label{planning section}
The differentiable nature of Log-GPIS permits straightforward integration with existing optimisation-based planners.
In this section, we present how Log-GPIS can be used together with the covariant Hamiltonian optimisation-based motion planner~(CHOMP)~\cite{Chomp} for obstacle avoidance planning.

Let $\mathbf{x}^{r} : t \mapsto \mathbf{x}^{r}(t)$ be the robot's trajectory. 
For simplicity, we consider the robot's position in the workspace over time, excluding orientation, so that $\mathbf{x}^{r}(t) \in \mathbb{R}^{D}$.
However, the framework can be easily extended to cases with orientation or with multiple joints. 
% In other words, the robot is modelled as a point mass without joints,  
% For simplicity, we consider a particular instance of CHOMP with the waypoint parametrisation, where the trajectory $\mathbf{x}$ is represented by a sequence of $M$ waypoints $\mathbf{x} = \{\mathbf{x}_{m}\} \subset \mathbb{R}^{D}, m  = 1 ... M$.

CHOMP minimises an objective \emph{functional} defined over a space of trajectories. 
We consider the conventional objective function as introduced in~\cite{Chomp} for smooth obstacle avoidance:
\begin{equation}\label{eq:chomp:objective}
    \mathcal{C}[\mathbf{x}^{r}] \equiv \int_{0}^{T} \frac{1}{2} || \dot{\mathbf{x}}^{r}(t) ||^{2} + \lambda c(\mathbf{x}^{r}(t)) || \dot{\mathbf{x}}^{r}(t) || dt.
\end{equation}
Here, the first term encourages smoothness by regularising the velocity. 
The second term is responsible for obstacle avoidance, which is enforced by the collision penalty term $c(\mathbf{x}^{r}(t))$ set as: 
\begin{equation}\label{eq:chomp:collision}
c(\mathbf{x}^{r}(t)) =
\begin{cases}
     -d(\mathbf{x}^{r}(t)) + \frac{1}{2}\epsilon, & \text{if } d(\mathbf{x}^{r}(t)) < 0 \\
     \frac{1}{2\epsilon}(d(\mathbf{x}^{r}(t)) - \epsilon)^{2}, & \text{if } 0 < d(\mathbf{x}^{r}(t)) \leq \epsilon \\
     0, & \text{otherwise.}
\end{cases}
\end{equation}

CHOMP incrementally minimises the objective functional $\mathcal{C}[\mathbf{x}^{r}]$ using its functional gradient\footnote{We focus only on gradient computation and omit the exact update process. Interested readers are referred to~\cite{Chomp}.}.
With the objective set as~\eqref{eq:chomp:objective}, the functional gradient is given by:
\begin{equation}\label{eq:chomp:grad-functional}
    \nabla \mathcal{C}[\mathbf{x}^{r}] = - \ddot{\mathbf{x}}^{r} + ||\dot{\mathbf{x}}^{r}|| \left( \left(I - \dot{\mathbf{x}}^{r}\dot{\mathbf{x}}^{rT} \right) \nabla c(\mathbf{x}^{r}) - c(\mathbf{x}^{r}) \kappa \right).
\end{equation}
Here, $\kappa = ||\dot{\mathbf{x}}^{r}||^{-2} \left(I - \dot{\mathbf{x}}^{r}\dot{\mathbf{x}}^{rT} \right) \ddot{\mathbf{x}}^{r} $ is the curvature vector. 
The gradient of the collision penalty term $c(\mathbf{x}^{r})$ is given by
\begin{equation}\label{eq:chomp:grad-collision}
    \nabla c(\mathbf{x}^{r}) = 
\begin{cases}
     -\nabla d(\mathbf{x}^{r}), & \text{if } d(\mathbf{x}^{r}) < 0 \\
     (d(\mathbf{x}^{r}) - \epsilon) \nabla d(\mathbf{x}^{r}) & \text{if } 0 < d(\mathbf{x}^{r}) \leq \epsilon \\
     0, & \text{otherwise.}    
\end{cases}    
\end{equation}

The merit of our framework is that we can use the Log-GPIS gradient inference equation~\eqref{eq:gradient} to efficiently and accurately compute the gradient~$\nabla d(\mathbf{x}^{r}) $, and, in turn, \eqref{eq:chomp:grad-functional},~\eqref{eq:chomp:grad-collision}.
This is because our formulation produces the gradients analytically, whereas conventional approaches rely on discrete grid-based approximation~\cite{Chomp,oleynikova2018loco}.

\section{Experimental Results}\label{results section}

In this section, we illustrate the performance of the proposed framework both qualitatively and quantitatively. 
% We consider three scenarios, comprising odometry only~\ref{subsec:odometry}, odometry and mapping~\ref{subsec:mapping}, and all of odometry, mapping and path planning~\ref{subsec:planning}. 
The framework is implemented in C++ and Matlab. 
All experiments were run on an 8-core i7 CPU at 2.5GHz.

% \begin{figure}[t]
% 	\centering
% 	\resizebox{1\linewidth}{!}{

% 	}
% 	\caption{50-run Monte Carlo simulation. For each of the box plots the error distribution with increasing Gaussian noise from $\sigma$ 0.01 to 0.3 is shown.}
% \vspace{-2ex}
% 	\label{RMSE_distance}
% \end{figure}

\begin{figure*}
  \centering
  \subfloat[]{\includegraphics[scale=0.25]{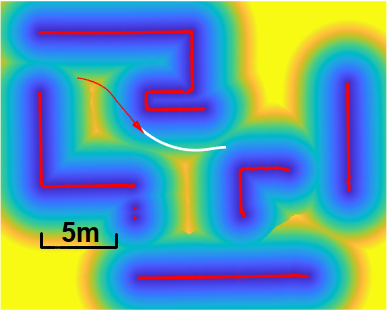}}
  \subfloat[]{\includegraphics[scale=0.25]{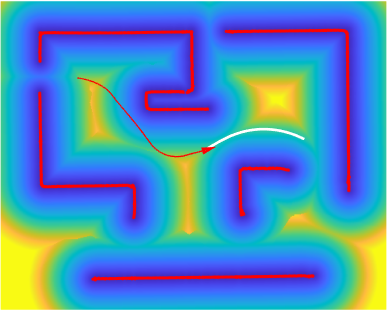}}
  \subfloat[]{\includegraphics[scale=0.25]{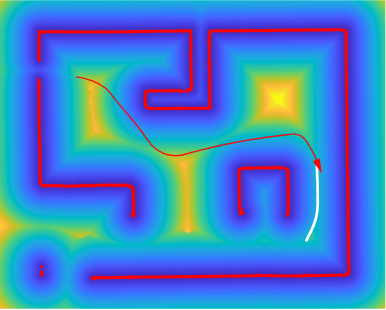}} 
  \subfloat[]{\includegraphics[scale=0.25]{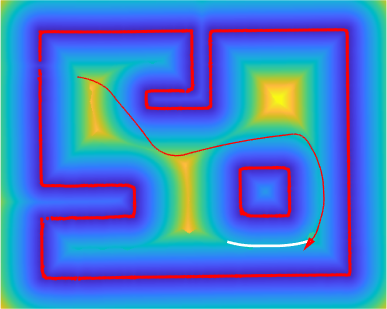}}
  \subfloat[]{\includegraphics[scale=0.25]{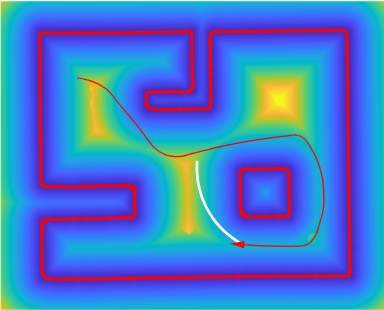}}
  \subfloat{\includegraphics[scale=0.25]{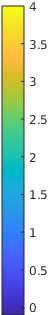}}
  \caption{The Log-GPIS-MOP on a simulated 2D dataset. The reconstructed surfaces are red lines. The colourmap represents the distance field varying from -4 to 4 meters. Red line with an arrow is the estimated trajectory where the robot travelled. The white line is the suggested path computed by the path planner using the global Log-GPIS.}
  \label{method_mapping_planning_odometry}
\end{figure*}

\edit{\subsection{Effects of Log Transformation}\label{subsec:planning}
In \cite{wu2021faithful}, we showed that our Log-GPIS has the advantage of both mapping and distance accurate estimation. In this section, we examine the effects of the log transformation on mapping, odometry and planning. 
We first illustrate the nominal behaviour of the full Log-GPIS-MOP framework. To do so, we first use a $20\times16$m simulated 2D LiDAR dataset from~\cite{Bhoram,occupancy2012gaussian}, the second simulated dataset in~\cite{soorajanilkumar}, and the real-world Intel Research Lab dataset~\cite{intel}.}
To emulate online planning with incremental mapping, we generate the plan from the robot's current pose to its future pose in the dataset, 50 time-steps ahead.
Although the robot does not strictly follow the generated plan, this serves as a useful proxy for performance in online scenarios.

The result for the first simulated dataset is shown in Fig.~\ref{method_mapping_planning_odometry}, with the travelled trajectory in red, the planned trajectory in white, and the underlying colourmap representing the EDF built so far.
Most notably, in Fig.~\ref{method_mapping_planning_odometry}c), it can be seen that the white planned trajectory avoids the obstacles even when they are not yet seen.
The same behaviour is observed in the more challenging scenario of the Intel Research Lab dataset~\cite{intel} illustrated in Fig.~\ref{fig:teaser}, where the white planned trajectories are smooth and stay a fixed distance away from the obstacles as intended.
This is because Log-GPIS extrapolates well to areas with missing data.
Moreover, in both cases, Log-GPIS-MOP produces accurate odometry and EDF, as was observed in Sec.~\ref{sec:simulation}.
We attribute such extrapolation to the log transformation, as it implicitly approximates the Eikonal equation across the entire space.

We now interrogate this hypothesis through comparison against GPIS-SDF~\cite{Bhoram} on the second simulated dataset.
For fairness, we implemented the same odometry and planning algorithms using GPIS-SDF.
In particular, we use the same weights for trajectory smoothness and collision avoidance when planning the trajectories. 

\begin{figure}
  \centering
  %\resizebox{\linewidth}{!}{
  \subfloat[\label{planning_comparison_a}]{\includegraphics[height=3.7cm]{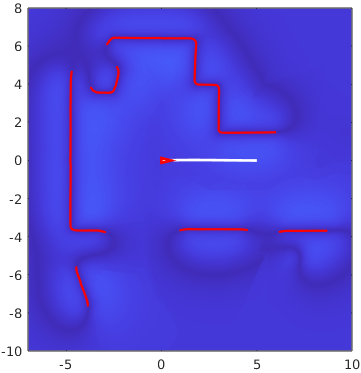}}
  \subfloat[\label{planning_comparison_b}]{\includegraphics[height=3.7cm]{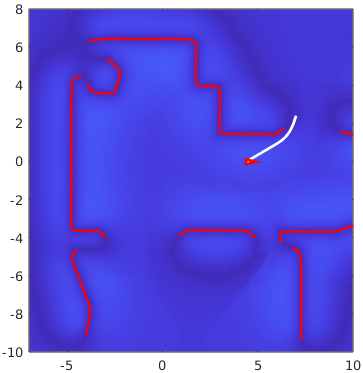}} 
  \subfloat{\includegraphics[height=3.5cm]{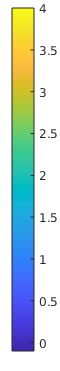}} \\
  \setcounter{subfigure}{2} 
  \subfloat[\label{planning_comparison_c}]{\includegraphics[height=3.7cm]{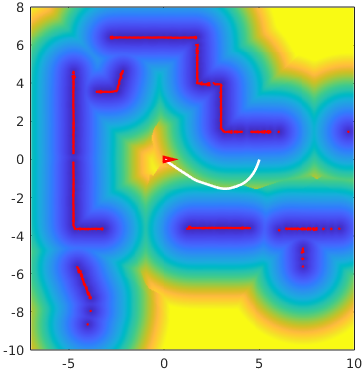}}
  \subfloat[\label{planning_comparison_d}]{\includegraphics[height=3.7cm]{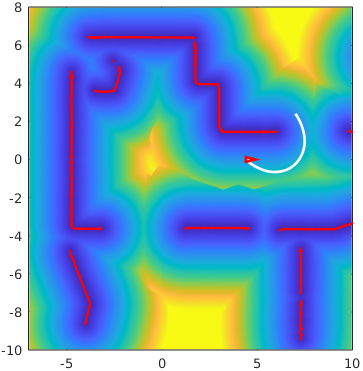}}
  \subfloat{\includegraphics[height=3.5cm]{figures/planning_comparison_colourmap.png}}
  %}
  \caption{Mapping and planning results of GPIS-SDF (a-b) and Log-GPIS-MOP (c-d). Red arrows show the robot's current pose. With the same collision avoidance weight, the planned trajectories (white) from Log-GPIS-MOP in c-d) are strongly biased towards the medial axis, whereas that of GPIS-SDF in a-b) are nearly straight lines, with a near-hit in b).}
  \label{planning_comparison}
\vspace{-2ex}
\end{figure}

The results are shown in Fig.~\ref{planning_comparison}.
We observed that odometry based on GPIS-SDF fails with a moderately large motion because GPIS-SDF produces increasingly inaccurate EDF predictions further away from previous measurements, unlike Log-GPIS.
Thus, we present a portion of the results before GPIS-SDF fails, in order to compare the planned trajectories.

The planned trajectories are shown in white in Fig.~\ref{planning_comparison}.
In Figs.~\ref{planning_comparison}(a-b), It can be seen that trajectory planning using GPIS-SDF results in nearly a straight line, with a near miss in Fig.~\ref{planning_comparison_b}.
On the other hand, with the same weighting for collision avoidance, Log-GPIS-MOP produces a plan strongly biased towards the medial axis of the environment (i.e. equidistant from obstacles).
%This is because GPIS-SDF predictions are inaccurately low further away from the measured surfaces.
These observations illustrate that the log transformation in Log-GPIS-MOP represents a critical improvement in prediction accuracy that is necessary for practical use in odometry and planning.

\begin{figure}
	\centering
	\subfloat[Trajectory comparison with different noise\label{simulation_tra:comparison}]{\includegraphics[height=0.18\textheight]{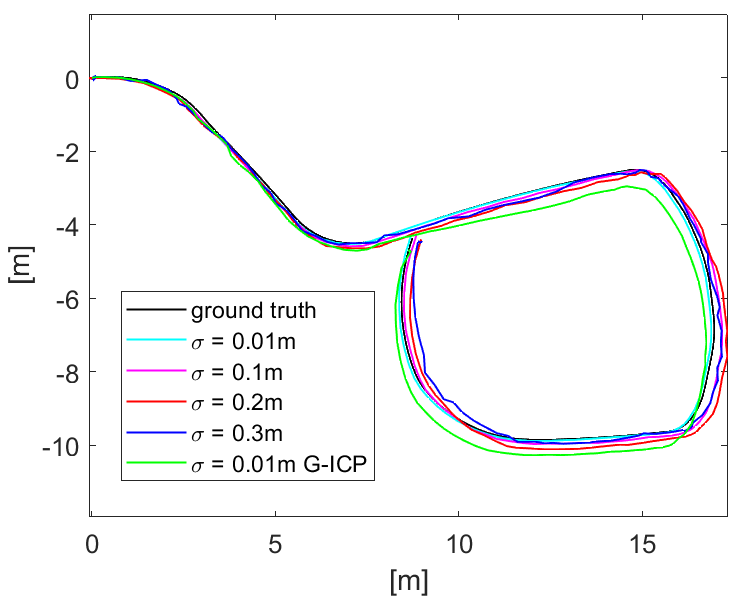}} \\
	\subfloat[Translation RMSE\label{simulation_tra:translation}]{\includegraphics[height=0.18\textheight]{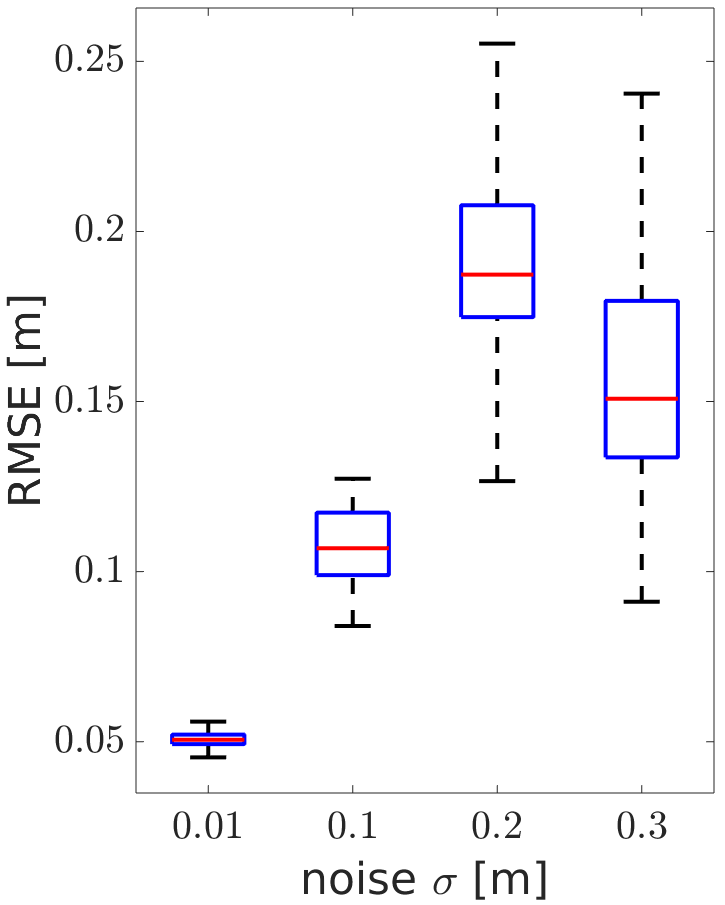}}
	\subfloat[Rotation RMSE\label{simulation_tra:rotation}]{\includegraphics[height=0.18\textheight]{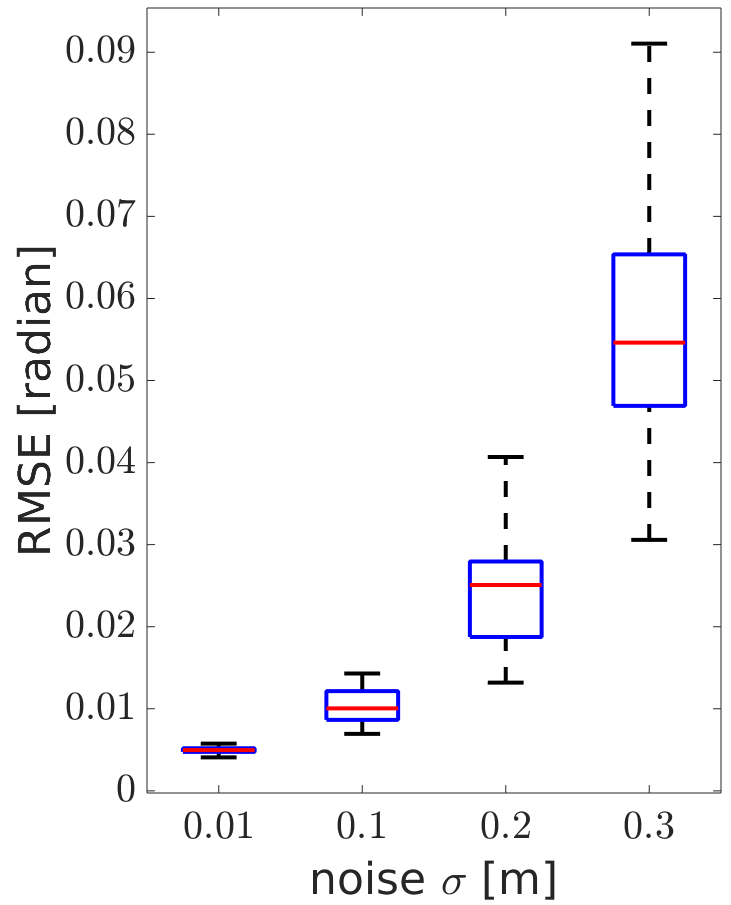}}
	\caption{Comparison of ground truth and estimated trajectories using Log-GPIS with varying levels of simulated measurement noise. The estimated trajectories in a) closely resemble the ground truth for all noise magnitudes. The translational and rotational RMSE in b) and c) exhibits acceptable increase with noise.}
	\label{simulation_tra}
\end{figure}

\begin{figure*}
  \centering
  \subfloat{\includegraphics[height=3cm]{figures/colorbar.png}}
  \setcounter{subfigure}{0}  
  \subfloat[Ground Truth\label{distance_simulation:ground_truth}]{\includegraphics[height=3cm]{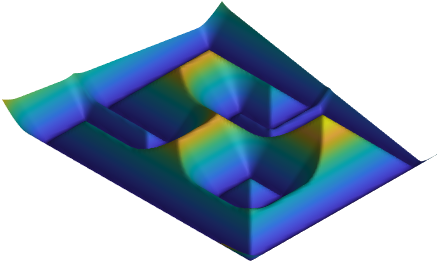}}
  \subfloat[Log-GPIS ($\sigma=0.01m$)\label{distance_simulation:loggpis}]{\includegraphics[height=3cm]{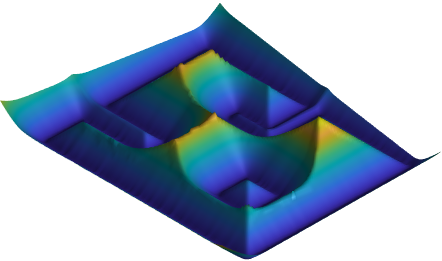}}
  \subfloat[RMSE\label{distance_simulation:rmse}]{\includegraphics[height=3.2cm]{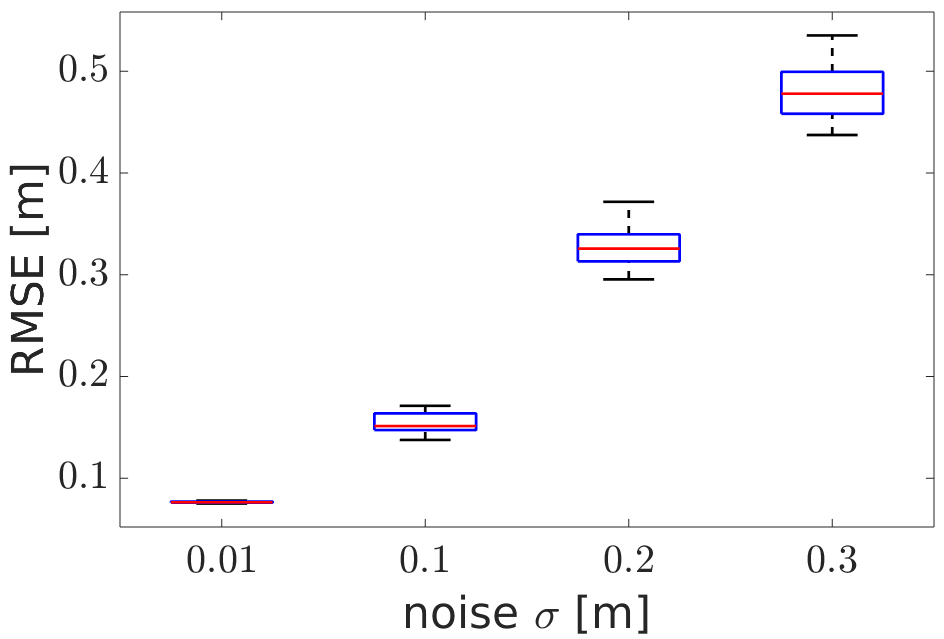}}
  \caption{Comparison of ground truth and estimated EDF with varying levels of noise. z-axes in b) and c) correspond to the distance value. 
  The estimated EDF with a realistic noise $\sigma=0.01m$ in c) is accurate and closely resembles the ground truth in b). The RMSE in d) exhibits a linear increase with noise.}
  \label{distance_simulation}
\end{figure*}

\subsection{Robustness to Noise}\label{sec:simulation}
We first illustrate the robustness of the proposed Log-GPIS-MOP framework against noise.
To do so, we use the same simulated 2D LiDAR dataset~\cite{Bhoram,occupancy2012gaussian} in Sec.~\ref{subsec:planning}, and add varying magnitudes of Gaussian noise from $0.01m$ to $0.3m$ \edit{to the sensor measurements}. \edit{Note that this dataset is captured by a Turtlebot with a Hokuyo range sensor in Gazebo. The original standard deviation of the measurement noise is $\sigma=0.01m$~\cite{Bhoram}. In this simulation, we only use the sensor measurements, the sensor poses from the sensor frame at each timestamp to the world frame are estimated with our framework.}

The result for odometry is shown in Fig.~\ref{simulation_tra:comparison}.
It can be seen that, with a realistic noise ($\sigma=0.01m$, cyan), the result is almost identical to the ground truth trajectory (black). 
Even with a significantly large noise of $\sigma=0.3m$ (blue), the estimated trajectory robustly has acceptable drift. \edit{For comparison, we plot the trajectory of Generalised-ICP (G-ICP)~\cite{segal2009generalized} in green. The odometry of G-ICP is frame-to-map and the sensor measurements are with Gaussian noise of $\sigma=0.01m$. As we can see, the green line drifts quickly even with the smallest noise.}
Figs.~\ref{simulation_tra:translation} and~\ref{simulation_tra:rotation} illustrate the root mean squared error (RMSE) over 50 Monte Carlo runs in translation and rotation respectively.
It can be seen that the overall error of Log-GPIS odometry remains small, demonstrating robustness against noise. 
In particular, even with a large sensor noise of $\sigma = 0.3m$, the maximum translation RMSE is less than 0.25m over a ~30m length trajectory.

In Fig.~\ref{distance_simulation}, we compare the reconstructed EDF~(Fig.~\ref{distance_simulation:loggpis}) against the ground truth~(Fig.~\ref{distance_simulation:ground_truth}). 
We visualise the EDF in 3D, with the z-axis being the distance field value.
The colourmap varies from blue to yellow, corresponding to the distance field value from 0 to 4 meters.
It can be seen that the EDF reconstructed using our method is close to identical to the ground truth, sharing the same peaks and valleys.

The RMSE plot in Fig.~\ref{distance_simulation:rmse} further shows that the reconstructed EDF is accurate across the entire space even with increasing noise values, with a median RMSE is 0.07629m given reasonable sensor noise of 0.01m. 
In particular, a linear trend is observed between the noise and the distance field RMSE.
This illustrates that the Log-GPIS technique can accurately predict the EDF far away from measurements even though the measurements are only available near the surface.

\subsection{Comparison to SLAM Frameworks}\label{sec:results:slam}
\begin{figure*}[ht]
	\centering
    \subfloat[Estimated Trajectory\label{museum_map:traj}]{\includegraphics[height=0.16\textheight]{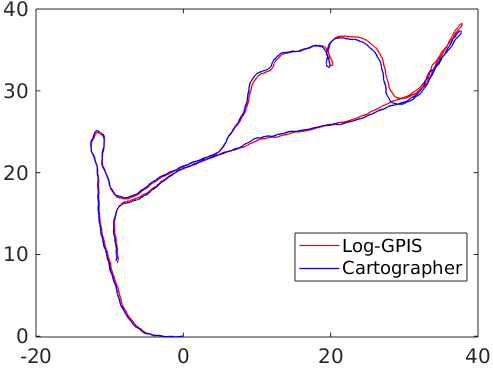}}
    \subfloat[Cartographer\label{museum_map:cartographer}]{\includegraphics[height=0.16\textheight]{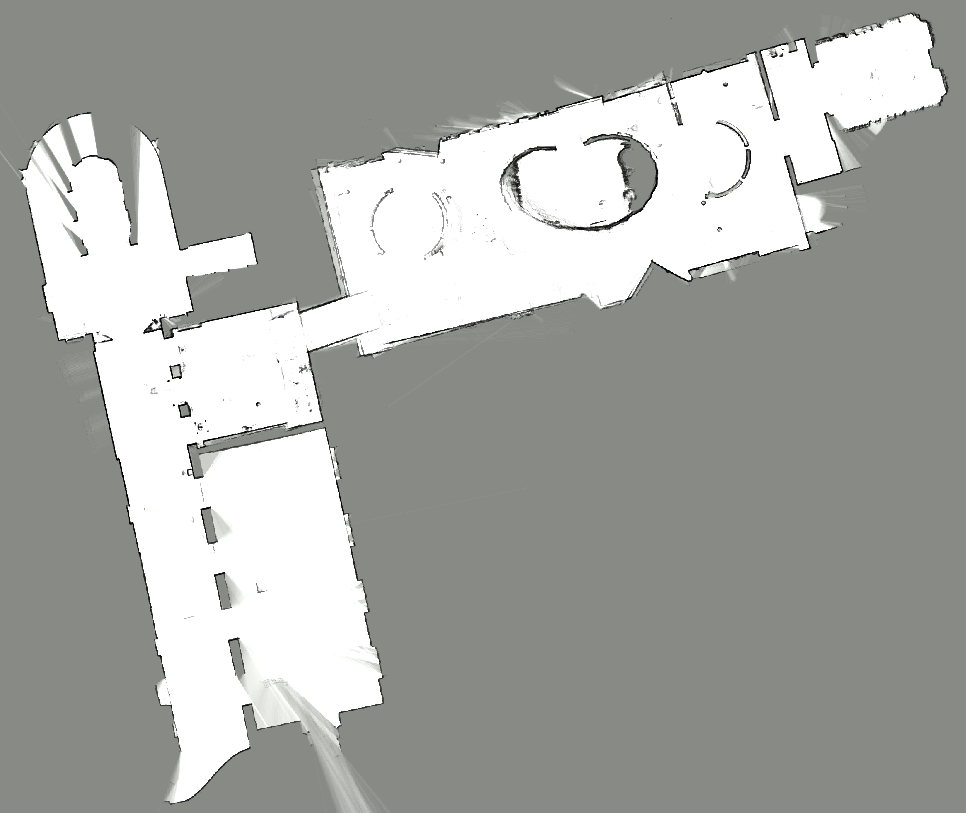}}
	\subfloat[Log-GPIS\label{museum_map:loggpis}]{\includegraphics[height=0.16\textheight]{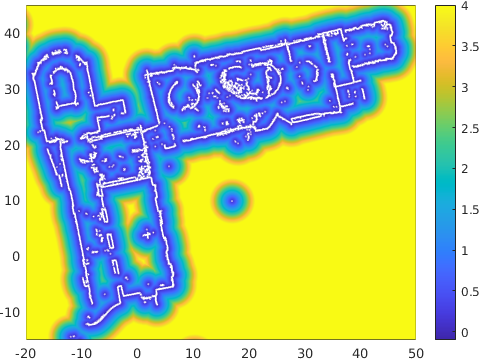}}
	\caption{a) Cartographer map using 2D LiDAR dataset. The resolution of the occupancy grid is 0.05 meters. The odometry and global optimisation are enabled. The rest of the parameters are set as default. b) As a comparison, the mapping result of our proposed method is shown. We can clearly see the wall and hallways. c) The surface modelling with distance field is demonstrated on the Intel dataset. The local minimum of iso-surfaces is in white colour.}
	\label{museum_map}
	\vspace{-2ex}
\end{figure*}
\begin{figure}
    \centering
	\includegraphics[width=0.55\columnwidth]{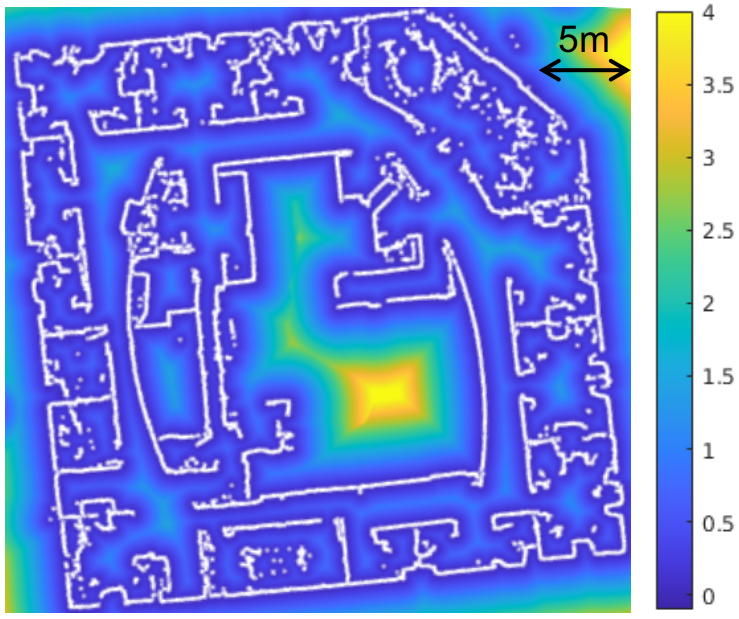}
    \caption{Reconstructed Log-GPIS map on the Intel Research Lab dataset.}
    \label{intel_map}
\end{figure}
\subsubsection{2D SLAM}
We compare the performance of Log-GPIS MOP for 2D SLAM against Cartographer~\cite{carto2016}, a popular open-source 2D SLAM framework. 
Cartographer produces an occupancy grid as opposed to an EDF produced by our framework.
We thus qualitatively examine the similarity of the produced maps.
Nonetheless, we quantitatively compare the odometry performance.
In doing so, Cartographer has the advantage that it additionally uses IMU readings for short-term odometry, as well as loop closure detection and pose-graph optimisation for long-term drift compensation.
Meanwhile, our incremental Log-GPIS odometry and mapping approach solely uses LiDAR measurements.
IMU fusion and loop closure remain avenues for future work. 

We use two real-world datasets for this comparison, namely the Deutsches Museum~\cite{carto2016} and the Intel Research Lab~\cite{intel} datasets.
The Deutsches Museum dataset was collected using a 2D LiDAR backpack.
% The dataset covers an area of $70m\times60m$, with a resolution of $0.1 m$. 
The Intel Research Lab dataset also consists of 2D LiDAR measurements (13631 frames) collected from a robot.
The dataset comprises multiple loops, with two loops in the corridor followed by detailed scans of each room.
% We focus on an area spanning $34m\times32m$, with a resolution of $0.1 m$.  
Since the Deutsches Museum dataset was collected in a relatively large real-world indoor environment, there is no ground truth trajectory.
On the other hand, we adopt the evaluation method suggested in~\cite{kummerle2009}, which computes relative poses between frames for the Intel Research Lab dataset, which we use to derive the ground truth poses.

The comparison using the Deutsches Museum dataset is shown in Fig.~\ref{museum_map}.
Due to the lack of ground truth poses, we qualitatively compare the trajectory from Log-GPIS-MOP against that from Cartographer~\cite{carto2016} as shown in Fig.~\ref{museum_map:traj}.
Even without explicit loop closure or IMU preintegration, our incremental Log-GPIS trajectory (red line) produces a close to identical result to Cartographer (blue line).
Similarly, we compare the mapping results in Figs.~\ref{museum_map:loggpis} and~\ref{museum_map:cartographer}.
In Fig.~\ref{museum_map:loggpis}, the colourmap shows the distance field values from 0 (blue) to 4 (yellow) meters. The reconstructed surface (i.e. minimum absolute distance) is shown in white.
We truncated the colour map at 4 meters for better visualisation of the building structures, although Log-GPIS remains accurate further away.
It can be seen that the reconstructed surface closely aligns with the occupancy map from Cartographer shown in Fig.~\ref{museum_map:cartographer}. 

% {\small \begin{table}[ht!]
% \caption{Quantitative comparison of errors on Intel Lab dataset}
% \begin{center}
% \begin{tabular}{|c|c|c|c|c|c|}
% \hline
%  & & SM  & GM & Cartographer & Log-GPIS\\
% \hline
% \textsl{Translation ($m$)} & \textsl{Mean} & 0.220 & 0.031 & 0.0229 & 0.0336 \\
% \hline 
% & \textsl{STD} & 0.296 & 0.026 & 0.0239 & 0.0253 \\
% \hline
% \textsl{Rotation ($^\circ$)} & \textsl{Mean} & 1.7  & 1.3 & 0.453 & 1.49\\
% \hline
% & \textsl{STD} & 4.8  & 4.7 & 1.335 & 4.55\\
% \hline 
% \end{tabular}
% \end{center}
% \label{2DTable}
% \end{table}}

{\small \begin{table}[ht!]
\caption{Quantitative comparison of errors on Intel Lab dataset}
\begin{center}
\begin{tabular}{|c|c|c|}
\hline
 & \textsl{Absolute translation[$m$]} & \textsl{Absolute rotation[$deg$]} \\
\hline
\textsl{Scan matching} & 0.220$\pm0.296$ & 1.7$\pm4.8$ \\
\hline 
\textsl{Log-GPIS} & 0.0336$\pm0.0253$ & 1.4904$\pm4.5507$ \\
\hline
\textsl{Graph Mapping} & 0.031$\pm0.026$ & 1.3$\pm4.7$ \\
\hline
\textsl{Cartographer} & 0.0229$\pm0.0239$ & 0.453$\pm1.335$ \\
\hline
\end{tabular}
\end{center}
\label{2DTable}
\end{table}}

We conduct a more quantitative analysis using the Intel Research Lab dataset~\cite{intel}. We present a comparison against the Scan Matching~\cite{scan_matching} (SM), Graph Mapping~\cite{graph_mapping} (GM) in addition to Cartographer, using the values reported in~\cite{carto2016}.Table.~\ref{2DTable} presents the absolute translational and rotational errors with standard deviation. 
Overall, Cartographer exhibits the lowest error owing to loop closure detection. 
Our framework outperforms the SM approach and demonstrates similar results to the GM and Cartographer approaches.
The mapping result in Fig.~\ref{intel_map} shows that the framework performs as expected, providing an accurate reconstruction of the wall surface, rooms and corridors.

{\small \begin{table}[t]
\caption{Quantitative comparison of errors with ORB-SLAM2 on a segment of the Teddy Bear dataset}
\begin{center}
\begin{tabular}{|c|c|c|}
\hline
 & ORB-SLAM2 & Log-GPIS   \\
\hline
\textsl{Translational error RMSE [m]} & 0.020317 & \textbf{0.017197} \\
\hline 
\textsl{Translational error mean [m]} & 0.017291 & \textbf{0.014583} \\
\hline
\textsl{Translational error median [m]} & 0.015385 & \textbf{0.013027} \\
\hline
\textsl{Translational error std [m]} & 0.010668 & \textbf{0.009114} \\
\hline
% \textsl{Translational error min [m]} & 0.000000 & 0.000000 \\
% \hline
\textsl{Translational error max [m]} & 0.050399 & \textbf{0.044593} \\
\hline
\textsl{Rotational error RMSE [deg]} & \textbf{1.225683} & 1.452126 \\
\hline 
\textsl{Rotational error mean [deg]} & \textbf{1.096656} & 1.289137 \\
\hline
\textsl{Rotational error median [deg]} & \textbf{0.018657} & 0.021269 \\
\hline
\textsl{Rotational error std [deg]} & \textbf{0.547399} & 0.668427 \\
\hline
% \textsl{Rotational error min [deg]} & 0.000000 & 0.000000 \\
% \hline
\textsl{Rotational error max [deg]} & \textbf{2.582371} & 3.734151 \\
\hline
\end{tabular}
\end{center}
\label{3DTable}
\end{table}}

\subsubsection{3D SLAM}\label{odometry real 3d section}
To evaluate the performance of Log-GPIS-MOP in the 3D setting, we compare it against ORB-SLAM2~\cite{paper:orbslam2}, a state-of-the-art feature-based SLAM framework. 
ORB-SLAM2 uses sparse keypoint features from the RGB images for odometry, whereas Log-GPIS-MOP does not use the RGB component. 
As ORB-SLAM2 does not produce a dense map, we only compare the odometry performance.
The quality of the estimated camera trajectory is evaluated using the tool from~\cite{sturm12iros_ws}.

We use a real-world RGB-D dataset called the Freiburg3 Teddy from TUM~\cite{sturm12iros_ws}, illustrated in Figs.~\ref{teddy_mesh:rgb_front} and~\ref{teddy_mesh:rgb_back}.
The dataset consists of RGB-D images collected using a Kinect sensor, along a ground-truth trajectory recorded using a motion-capture system with eight tracking cameras.
The image resolution is $640 \times 480$ at a frequency of $30$ Hz. 
We use the first 100 frames for evaluation.

The result is shown in Table.~\ref{3DTable}.
As can be seen, Log-GPIS-MOP outperforms ORB-SLAM in translational error, although ORB-SLAM2 shows lower rotational error.
In other words, Log-GPIS-MOP exhibits comparable performance to ORB-SLAM2, even though it does not use RGB features for alignment as ORB-SLAM2 does. 

\subsection{Comparison to Distance Field Mapping Frameworks}

\begin{figure}[ht]
	\centering
	\subfloat[Image of Teddy's front\label{teddy_mesh:rgb_front}]{\includegraphics[height=4cm]{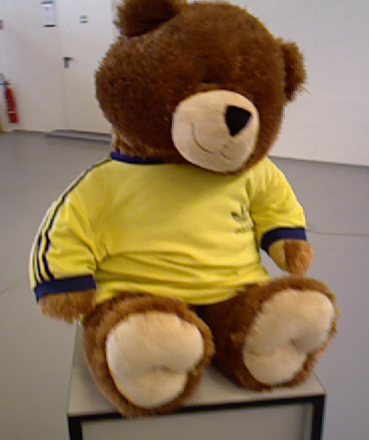}}
	\subfloat[Image of Teddy's back\label{teddy_mesh:rgb_back}]{\includegraphics[height=4cm]{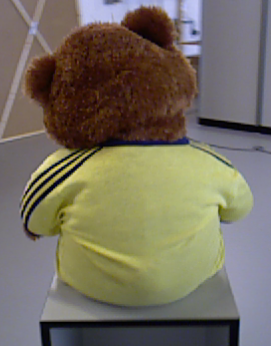}} \\
	\subfloat[Voxblox + GT trajectory (front)\label{teddy_mesh:voxblox_front}]{\includegraphics[height=4cm]{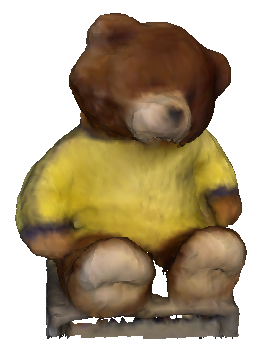}}
	\subfloat[Voxblox + GT trajectory (back)\label{teddy_mesh:voxblox_back}]{\includegraphics[height=4cm]{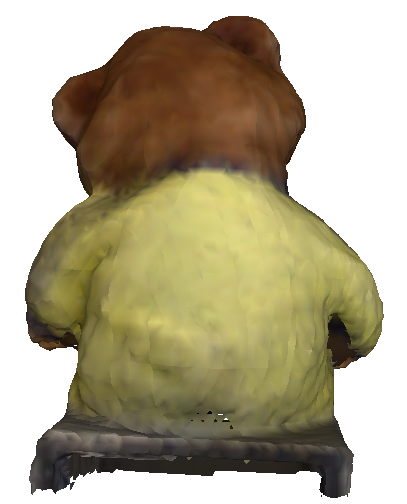}} \\
	\subfloat[Log-GPIS-MOP (front)\label{teddy_mesh:loggpis_front}]{\includegraphics[height=4.15cm]{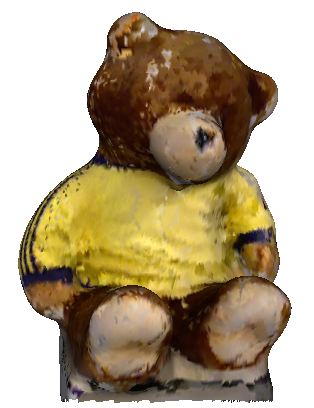}}
	\subfloat[Log-GPIS-MOP (back)\label{teddy_mesh:loggpis_back}]{\includegraphics[height=4.05cm]{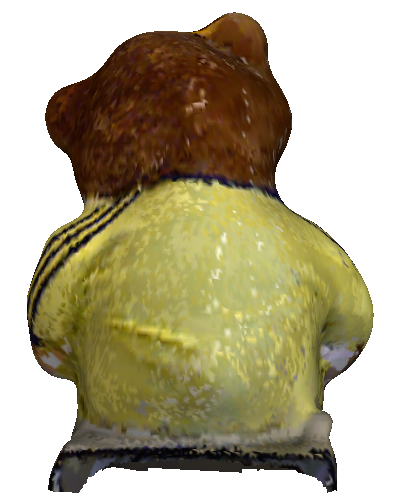}}
% 	\subfloat[Log-GPIS-SLAM mesh]{\includegraphics[height=4cm]{figures/smoothness.png}}
	\caption{Qualitative evaluation on Freiburg3 Teddy dataset. a) and b) show the raw images roughly at the same angle as the reconstructed mesh. Voxblox uses the ground truth poses for the reconstruction. In contrast, our method has no prior for the localisation. }
	\label{teddy mesh}
	\vspace{-2ex}
\end{figure}

We compare the performance of Log-GPIS-MOP in surface reconstruction against Voxblox~\cite{Voxblox}, a state-of-the-art distance field mapping and surface reconstruction framework.
Since Voxblox is a mapping-only framework that requires external odometry, we allow Voxblox to use the \emph{ground truth} trajectory from the dataset. 
Given the point cloud data and ground truth trajectory, Voxblox generates similar outputs as ours including the surface mesh, and the EDF. 
This is achieved by building a TSDF first, followed by wavefront propagation from some initial voxels for a certain distance to compute the EDF values. 
A dense mesh is reconstructed from the TSDF.
The resolution and TSDF are set at 0.01m and 0.03m respectively.
For a fair comparison, Log-GPIS-MOP uses a grid with the same resolution as testing points.
The rest of the parameters of Voxblox are set as default. 

We use two datasets, the Freiburg3 Teddy~\cite{sturm12iros_ws} used in Sec.~\ref{sec:results:slam} and the Cow and Lady dataset~\footnote{https://projects.asl.ethz.ch/datasets/doku.php?id=iros2017}.
For this comparison, we use the first 600 frames of the dataset, which corresponds to the first loop around the teddy bear.

\begin{figure*}[t]
  \centering
  %\setlength\tabcolsep{1pt}
  %\begin{tikzpicture}
       \resizebox{0.9\linewidth}{!}{
  \subfloat[Raw image of the scene\label{cow:scene_photo}]
  {\includegraphics[height=3.5cm]{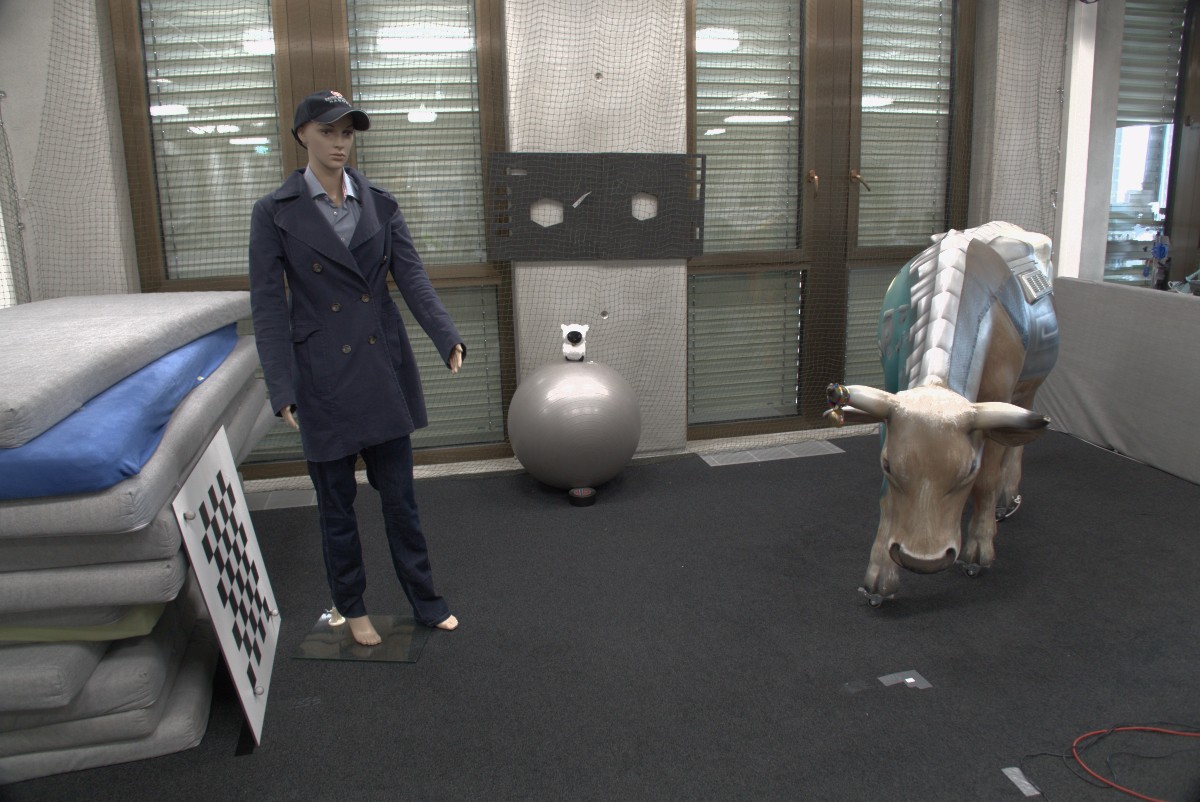}} 
  
  \subfloat[Log-GPIS mesh\label{cow:log-gpis}]
  {\includegraphics[height=3.5cm]{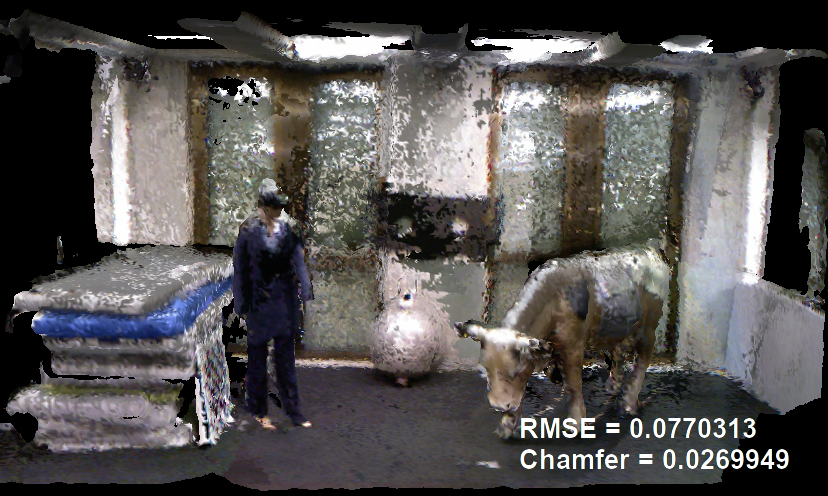}}
  %\draw [violet, thick] (3,3) node[right, blue,scale=0.75] {hahaha};

  \subfloat[Voxblox mesh\label{cow:voxblox}]{\includegraphics[height=3.5cm]{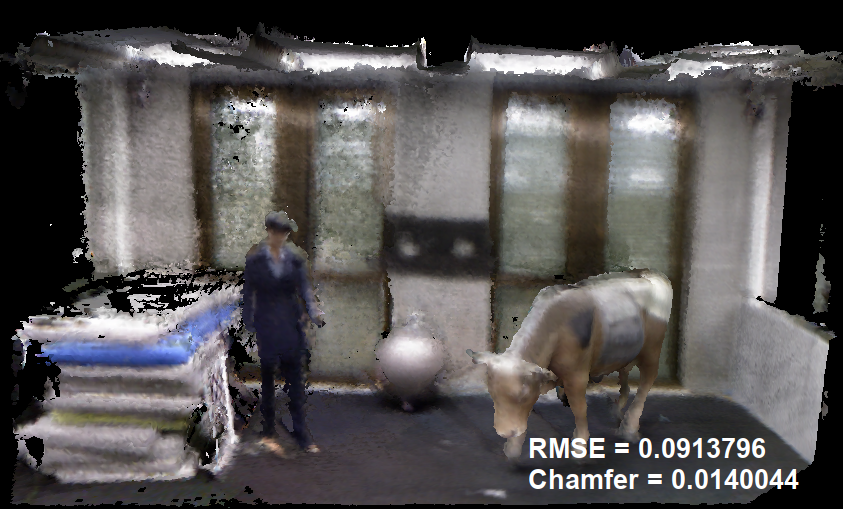}}}
  %\end{tikzpicture}
  
  \caption{3D odometry and mapping evaluation on the cow and lady dataset. a) shows the raw image of the scene. b) and c) show the reconstructed meshes of our framework and Voxblox respectively. Note that Voxblox fuses the colour for the final reconstruction. In contrast, we directly take the colour from the global merged point cloud. \edit{b) and c) have the RMSE and Chamfer distance values on the right bottom as the comparison of reconstruction quality.}}
  \label{ladycow_mesh}
\vspace{-2ex}
\end{figure*}

The Cow and Lady dataset includes fibreglass models of a large cow and a lady standing side by side in a room, as illustrated in Fig.~\ref{cow:scene_photo}.
The dataset consists of RGB-D point clouds collected using a Kinect RGB-D camera, with camera trajectory captured using a Vicon motion system. 
We use the first 450 frames of the dataset. 
This dataset is much more challenging than the Freiburg3 Teddy dataset, as the camera motion is relatively fast, with large changes in orientation. 
Furthermore, the Vicon trajectory is occasionally misaligned.
We use Log-GPIS-MOP to refine such misalignment by using the Vicon trajectory as a prior for odometry. 

The results for the Freiburg3 Teddy dataset are shown in Fig.~\ref{teddy mesh}.
Figs.~\ref{teddy mesh} c-d) and e-f) show the reconstructed map using Voxblox and Log-GPIS-MOP respectively.
Most notably, it can be seen that there is a gap in the Voxblox result (Figs.~\ref{teddy mesh} c-d)) between the hind paws due to missing data, whereas no such gap exists in the Log-GPIS-MOP result (Figs.~\ref{teddy mesh} e-f)).
This is because Log-GPIS-MOP naturally allows dealing with incomplete and sparse data, as GPIS allows extrapolation at unseen points.

Qualitatively, our reconstruction is closer to the image than Voxblox, especially the head and hind paws. 
Some colour differences are apparent due to light source variations. 
Voxblox fuses the colour of each frame. 
In contrast, our method obtains the colour directly from the global merged point cloud, which clearly produces the three black stripes on the sleeve of the yellow shirt.

\begin{figure}[t]
  \centering
  %\resizebox{\linewidth}{!}{
  \includegraphics[height=4.5cm]{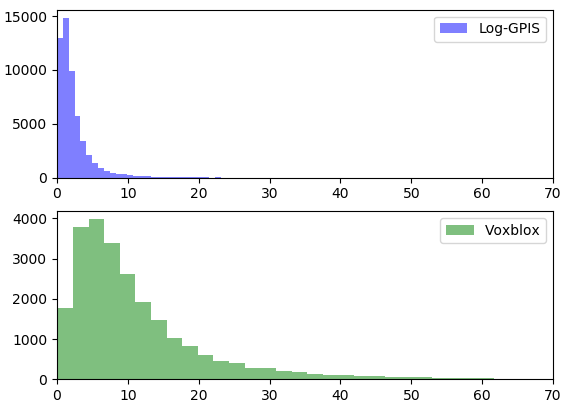}
  %}
  \caption{A histogram of the angle distribution of the normals in degrees. The top figure is Log-GPIS and the bottom figure is Voxblox.}
  \label{smoothness}
\vspace{-2ex}
\end{figure}

%The histogram of our Log-GPIS and Voxblox shows that our reconstructed normals are smoother than Voxblox. 

It is challenging to quantitatively evaluate the surface reconstruction performance since no ground truth is available for the Teddy bear's geometry.
\edit{Instead, we examine the surface smoothness by computing the difference between the estimated normal vectors and the smoothed normal over the ten nearest neighbours.
Fig.~\ref{smoothness} shows that Log-GPIS-MOP produces a smoother surface since GPIS has the ability to filter out noisy measurements.}

% Instead, we examine the precision by computing the difference between the estimated normal vectors and the smoothed normal over ten nearest neighbours, as this reveals how noisy the estimated normal vectors are. 
% In Fig.~\ref{smoothness}, it can be seen that the normal vectors from Log-GPIS-MOP are less noisy than that of Voxblox.
% This implies that Log-GPIS-MOP produces a smoother surface with less noisy normal vectors. 

The reconstruction results on the Cow and Lady dataset using Log-GPIS-MOP and Voxblox are shown in Figs.~\ref{ladycow_mesh}.
It can be seen that there is a gap at the top of the mattresses on the left-hand side in the Voxblox result due to missing data, whereas Log-GPIS-MOP does not have the same gap.
This is because Log-GPIS-MOP has the advantage of extrapolating the gaps in data, as we saw in the Freiburg3 Teddy dataset. Fig.~\ref{cow:scene_photo} shows the raw RGB image of the scene. 
Log-GPIS and Voxblox both produce similar reconstructed meshes shown in Fig.~\ref{cow:log-gpis} and Fig.~\ref{cow:voxblox} respectively. \edit{To compare the reconstructed surfaces quantitatively, we use the RMSE and Chamfer distance metrics against Voxblox. As demonstrated in Fig.~\ref{cow:log-gpis}, our surface quality has comparable results as Voxblox in Fig.~\ref{cow:voxblox}. }

\begin{figure*}[t]
  \centering
  \resizebox{\linewidth}{!}{
  \subfloat[Ground truth distance]{\includegraphics[height=3.5cm]{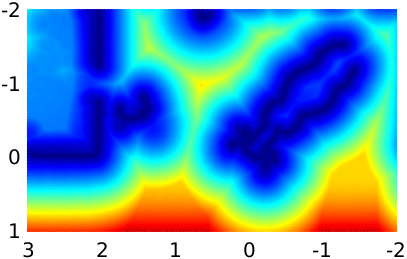}}
  \subfloat[Log-GPIS distance\label{cow:distance_log_gpis}]{\includegraphics[height=3.5cm]{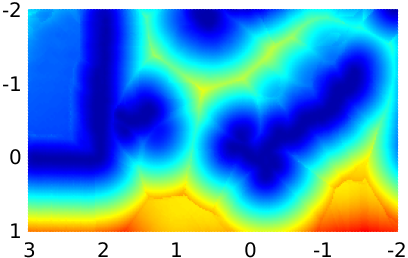}}
  \subfloat[Voxblox distance\label{cow:distance_voxblox}]{\includegraphics[height=3.5cm]{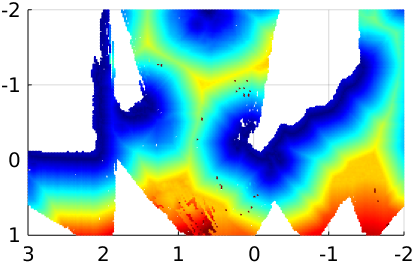}}
  \setcounter{subfigure}{2}
  \subfloat{\includegraphics[height=3.5cm]{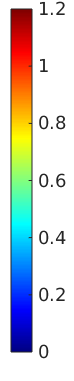}}
  \setcounter{subfigure}{3}
  \subfloat[\label{cow:distance_rmse}]{\includegraphics[height=3.5cm]{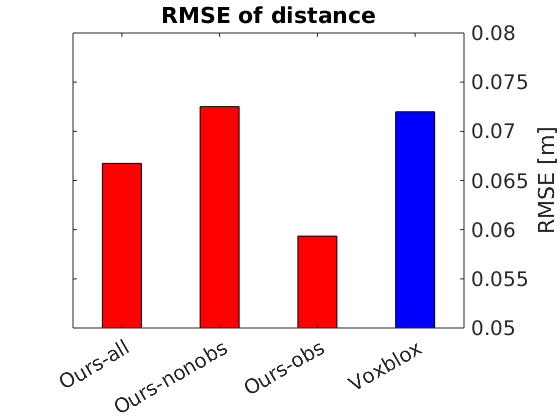}}
  }
  \caption{A horizontal slice ($5m\times3m$) shows the distance accuracy given Log-GPIS odometry and mapping on the cow and lady dataset. The slice is $0.8$m above the ground. \edit{a), b), and c) show from the top view \third{with scales in meters}. For the distance RMSE in d), note that Voxblox estimates the distance for the observed area only. Our method is able to predict not only in the observed area (Ours-obs) but also extrapolate the distance field to un-observed regions (Ours-nonobs). The RMSE shows that our method outperforms Voxblox, in the observed regions, which is the one that is fair to compare with Voxblox. It also shows that the prediction of the unobserved area from our method has similar errors than Voxblox in observed areas.}}
  \label{ladycow_distance}
\vspace{-2ex}
\end{figure*}

To further examine the behaviour of the two frameworks, we compare 2D slices of the ground-truth and estimated distance fields in Fig.~\ref{ladycow_distance}.
We choose a horizontal slice $0.8$m above the ground.
As shown in Fig.~\ref{cow:distance_voxblox}, Voxblox only computes the EDF within the sensor's field of view, whereas Log-GPIS-MOP naturally predicts the EDF value at all points as can be seen in Fig.~\ref{cow:distance_log_gpis}.
For a quantitative evaluation, we compare against a ground-truth distance field computed using the global point cloud from a Leica scanner.
We compute the RMSE for Voxblox only within the sensor field of view, as Voxblox does not compute the EDF outside this region.
Fig.~\ref{cow:distance_rmse} shows that \edit{our RMSE for the full region (Ours-all) and our prediction in observed regions (Ours-obs) clearly outperform Voxblox. The RMSE of the unobserved area 
(Ours-nonobs) has a similar performance to Voxblox in the observed regions.} %Thus, Log-GPIS-MOP outperforms Voxblox in both accuracy of the EDF and the prediction range.

%the RMSE error of Log-GPIS and Voxblox are 0.0672m and 0.0735m respectively. 

Regarding the computation time, the full Log-GPIS-MOP, using for example the simulated dataset~\cite{soorajanilkumar} (angular range from $0^\circ$ to $360^\circ$ with $1^\circ$ resolution) consumes a median computational time of 3.63s. Individually, the frame-to-map odometry consumes 1.98s. Incremental mapping takes 0.03s and path planning uses 2.29s. 
\edit{The final map reconstruction including post-processing takes an additional 29.12s with 190376 querying points. For the 3D odometry and mapping on the cow and lady dataset, the median computational time for each frame is 20.38s, individually 16.59s for the frame-to-map odometry, and 3.79s for incremental mapping. We use 13517926 querying points to perform marching cube. The final dense reconstruction takes an additional 26.94 minutes. When the framework runs incrementally, the odometry is the most computationally expensive due to each frame requiring inverting the covariance matrix of the size of the training points. In the future, we  will investigate the use of inducing points to make it more efficient.}

\edit{To summarise, Log-GPIS-MOP has been evaluated qualitatively and quantitatively on both simulated datasets~\cite{Bhoram,occupancy2012gaussian} and public real-word datasets~\cite{intel,carto2016,sturm12iros_ws,Voxblox}. Based on Log-GPIS representation, Log-GPIS-MOP provides competitive results against state-of-the-art frameworks in odometry, surface reconstruction, and obstacle avoidance. One limitation of our proposed odometry is the difficulty of finding the correct alignment in scenes lacking curvatures. This is because the distance error is the same along flat surfaces with no features. Another limitation that is present in most scan matching-based approaches is that the difference between the consecutive observations has to be relatively small in order to make sure there is sufficient overlap between consecutive measurements. Time complexity is also an issue in particular for odometry estimation that we are aiming to address in future work.

}

% \begin{figure}[t]
%   	\centering
% 	\resizebox{\linewidth}{!}{
% 	\includegraphics[]{figures/planning_comparison2.png}
% 	}
% 	\caption{Mapping and planning results of GPIS-SDF (a-b) and Log-GPIS-MOP (c-d). Red arrows show the robot's current pose. With the same collision avoidance weight, the planned trajectories (white) from Log-GPIS-MOP in c-d) are strongly biased towards the medial axis, whereas that of GPIS-SDF in a-b) are nearly straight lines, with a near-hit in b).
% 	}
%     \vspace{-2ex}
% 	\label{planning_comparison}
% \end{figure}

\section{Conclusion}\label{conclusion section}
We proposed Log-GPIS-MOP, a unified probabilistic framework for mapping, odometry and planning based on Log-GPIS. 
The main ingredient is the Log-GPIS representation, which allows accurate prediction of the EDF and its gradients.
By exploiting the global and local Log-GPIS, we presented a sequential odometry formulation for the incremental mapping and planning approaches, and a batch optimisation as post-processing to refine a sequence of poses. 
For Log-GPIS mapping, two different pipelines were proposed: the incremental mapping that appends the Log-GPIS with incoming point clouds and the post-processing mapping, which simply uses the output of the odometry and the raw point clouds altogether to recover a global GPIS. 
Concurrently, a path planning approach uses the reconstructed map to compute an optimal collision-free trajectory in the environment. 
Extensive analysis has been conducted to evaluate the proposed method on simulated and real datasets in both 2D and 3D against state-of-the-art frameworks. 
Our experiments showed that Log-GPIS-MOP exhibits comparable results to the state-of-the-art frameworks for localisation, surface mapping and obstacle avoidance, even without using IMU data or loop closure detection.
Future work lies in loop closure detection to develop a full SLAM solution that will improve long-term robustness in large-scale environments. \edit{Combining the dynamic module from~\cite{liu2021active} with the proposed Log-GPIS-MOP is an interesting and promising avenue for future work that can expand the applications to more complicated scenarios.} 
\edit{The predictive variance can be used in interesting planning problems such as information gathering~\cite{masha_info_gathering} or planning under uncertainty~\cite{brian_MIUCB}.}
Further, we would like to recover the sign of EDF and develop an efficient implementation that will allow online operation. %\edit{Inspired by the neural-based representation, we could combine the MLP with GP as the new post-processing mapping for better colour rendering.}

\bibliographystyle{IEEEtran}
\bibliography{reference}

\begin{thebibliography}{10}
\providecommand{\url}[1]{#1}
\csname url@samestyle\endcsname
\providecommand{\newblock}{\relax}
\providecommand{\bibinfo}[2]{#2}
\providecommand{\BIBentrySTDinterwordspacing}{\spaceskip=0pt\relax}
\providecommand{\BIBentryALTinterwordstretchfactor}{4}
\providecommand{\BIBentryALTinterwordspacing}{\spaceskip=\fontdimen2\font plus
\BIBentryALTinterwordstretchfactor\fontdimen3\font minus
  \fontdimen4\font\relax}
\providecommand{\BIBforeignlanguage}[2]{{%
\expandafter\ifx\csname l@#1\endcsname\relax
\typeout{** WARNING: IEEEtran.bst: No hyphenation pattern has been}%
\typeout{** loaded for the language `#1'. Using the pattern for}%
\typeout{** the default language instead.}%
\else
\language=\csname l@#1\endcsname
\fi
#2}}
\providecommand{\BIBdecl}{\relax}
\BIBdecl

\bibitem{Cadena16tro-SLAMfuture}
C.~Cadena, L.~Carlone, H.~Carrillo, Y.~Latif, D.~Scaramuzza, J.~Neira, I.~Reid,
  and J.~Leonard, ``Past, present, and future of simultaneous localization and
  mapping: Towards the robust-perception age,'' \emph{Trans. on Rob.}, p.
  1309–1332, 2016.

\bibitem{intel}
\BIBentryALTinterwordspacing
D.~Haehnel, \emph{Cyrill Stachniss—Robotics Datasets}, Intel Lab, 2019.
  [Online]. Available:
  \url{http://www2.informatik.uni-freiburg.de/~stachnis/datasets/}
\BIBentrySTDinterwordspacing

\bibitem{paper:orbslam2}
R.~Mur{-}Artal and J.~D. Tard{\'{o}}s, ``{ORB-SLAM2:} an open-source {SLAM}
  system for monocular, stereo, and {RGB-D} cameras,'' \emph{IEEE Transactions
  on Robotics (T-RO)}, vol.~33, no.~5, pp. 1255--1262, 2017.

\bibitem{thoma2019mapping}
J.~Thoma, D.~P. Paudel, A.~Chhatkuli, T.~Probst, and L.~V. Gool, ``Mapping,
  localization and path planning for image-based navigation using visual
  features and map,'' in \emph{Proc. of CVPR}, 2019, pp. 7383--7391.

\bibitem{vins-mono}
T.~Qin, P.~Li, and S.~Shen, ``Vins-mono: A robust and versatile monocular
  visual-inertial state estimator,'' \emph{IEEE Transactions on Robotics},
  2018.

\bibitem{kinectfusion}
S.~Izadi, D.~Kim, O.~Hilliges, D.~Molyneaux, R.~Newcombe, P.~Kohli, J.~Shotton,
  S.~Hodges, D.~Freeman, A.~Davison \emph{et~al.}, ``Kinectfusion: real-time 3d
  reconstruction and interaction using a moving depth camera,'' in \emph{Proc.
  of the ACM symposium}, 2011, pp. 559--568.

\bibitem{Chomp}
M.~Zucker, N.~Ratliff, A.~D. Dragan, M.~Pivtoraiko, M.~Klingensmith, C.~M.
  Dellin, J.~A. Bagnell, and S.~S. Srinivasa, ``Chomp: Covariant hamiltonian
  optimization for motion planning,'' \emph{Int. J. of Rob. Res.(IJRR)}, pp.
  1164--1193, 2013.

\bibitem{reijgwart2020voxgraph}
V.~{Reijgwart}, A.~{Millane}, H.~{Oleynikova}, R.~{Siegwart}, C.~{Cadena}, and
  J.~{Nieto}, ``Voxgraph: Globally consistent, volumetric mapping using signed
  distance function submaps,'' \emph{IEEE Robotics and Automation Letters},
  2020.

\bibitem{Microsoft}
O.~Williams and A.~Fitzgibbon, ``Gaussian process implicit surfaces,'' 2007.

\bibitem{LanRAL20}
L.~Wu, R.~Falque, V.~Perez-Puchalt, L.~Liu, N.~Pietroni, and T.~Vidal-Calleja,
  ``Skeleton-based conditionally independent gaussian process implicit surfaces
  for fusion in sparse to dense 3d reconstruction,'' \emph{IEEE Robotics and
  Automation Letters}, no.~2, pp. 1532--1539, 2020.

\bibitem{paper:GeomPrior}
W.~{Martens}, Y.~{Poffet}, P.~R. {Soria}, R.~{Fitch}, and S.~{Sukkarieh},
  ``Geometric priors for gaussian process implicit surfaces,'' \emph{IEEE
  Robotics and Automation Letters (RA-L)}, pp. 373--380, 2017.

\bibitem{Bhoram}
B.~Lee, C.~Zhang, Z.~Huang, and D.~D. Lee, ``Online continuous mapping using
  gaussian process implicit surfaces,'' in \emph{2019 International Conference
  on Robotics and Automation (ICRA)}, 2019.

\bibitem{stork2020ensemble}
J.~A. Stork and T.~Stoyanov, ``Ensemble of sparse gaussian process experts for
  implicit surface mapping with streaming data,'' \emph{2020 IEEE International
  Conference on Robotics and Automation(ICRA)}, 2020.

\bibitem{wu2021faithful}
L.~Wu, K.~M.~B. Lee, L.~Liu, and T.~Vidal-Calleja, ``Faithful euclidean
  distance field from log-gaussian process implicit surfaces,'' \emph{IEEE
  Robotics and Automation Letters}, pp. 2461--2468, 2021.

\bibitem{Crane}
K.~Crane, C.~Weischedel, and M.~Wardetzky, ``Geodesics in heat,'' \emph{ACM
  Transactions on Graphics}, 2012.

\bibitem{schneider2018maplab}
T.~Schneider, M.~T. Dymczyk, M.~Fehr, K.~Egger, S.~Lynen, I.~Gilitschenski, and
  R.~Siegwart, ``maplab: An open framework for research in visual-inertial
  mapping and localization,'' \emph{IEEE Robotics and Automation Letters},
  2018.

\bibitem{MVE}
S.~Fuhrmann, F.~Langguth, and M.~Goesele, ``Mve: A multi-view reconstruction
  environment,'' in \emph{Eurographics Workshop on Graphics and Cultural
  Heritage (GCH)}, 2014, pp. 11--18.

\bibitem{tsdf}
B.~Curless and M.~Levoy, ``A volumetric method for building complex models from
  range images,'' in \emph{Proc. of the Conference on Computer Graphics and
  Interactive Techniques}, 1996, p. 303–312.

\bibitem{ICP}
P.~J. Besl and N.~D. McKay, ``Method for registration of 3-d shapes,'' in
  \emph{Sensor fusion IV: control paradigms and data structures}.\hskip 1em
  plus 0.5em minus 0.4em\relax International Society for Optics and Photonics,
  1992, pp. 586--606.

\bibitem{rgbd-mapping}
P.~Henry, M.~Krainin, E.~Herbst, X.~Ren, and D.~Fox, ``Rgb-d mapping: Using
  kinect-style depth cameras for dense 3d modeling of indoor environments,''
  \emph{Int. J. of Rob. Res.(IJRR)}, pp. 647--663, 2012.

\bibitem{Whelan2015ElasticFusionDS}
T.~Whelan, S.~Leutenegger, R.~F. Salas-Moreno, B.~Glocker, and A.~J. Davison,
  ``Elasticfusion: Dense slam without a pose graph,'' in \emph{Robotics:
  Science and Systems}, 2015.

\bibitem{zobeidi2020semanticGP}
E.~Zobeidi, A.~Koppel, and N.~Atanasov, ``Dense incremental metric-semantic
  mapping via sparse gaussian process regression,'' in \emph{2020 IEEE/RSJ
  International Conference on Intelligent Robots and Systems (IROS)}.\hskip 1em
  plus 0.5em minus 0.4em\relax IEEE, 2020, pp. 6180--6187.

\bibitem{rosinol2020kimera}
A.~Rosinol, M.~Abate, Y.~Chang, and L.~Carlone, ``Kimera: an open-source
  library for real-time metric-semantic localization and mapping,'' in
  \emph{2020 IEEE International Conference on Robotics and Automation
  (ICRA)}.\hskip 1em plus 0.5em minus 0.4em\relax IEEE, 2020, pp. 1689--1696.

\bibitem{vasilopoulos2020semantic_planning}
V.~Vasilopoulos, G.~Pavlakos, S.~L. Bowman, J.~D. Caporale, K.~Daniilidis,
  G.~J. Pappas, and D.~E. Koditschek, ``Reactive semantic planning in
  unexplored semantic environments using deep perceptual feedback,'' \emph{IEEE
  Robotics and Automation Letters}, vol.~5, no.~3, pp. 4455--4462, 2020.

\bibitem{occupancy}
A.~Elfes, ``Using occupancy grids for mobile robot perception and navigation,''
  \emph{Computer}, vol.~22, no.~6, pp. 46--57, 1989.

\bibitem{Octomap}
A.~Hornung, K.~M. Wurm, M.~Bennewitz, C.~Stachniss, and W.~Burgard, ``Octomap:
  an efficient probabilistic 3d mapping framework based on octrees,''
  \emph{Autonomous Robots}, pp. 189--206, 2013.

\bibitem{carto2016}
W.~Hess, D.~Kohler, H.~Rapp, and D.~Andor, ``Real-time loop closure in 2d lidar
  slam,'' in \emph{2016 IEEE International Conference on Robotics and
  Automation (ICRA)}, 2016, pp. 1271--1278.

\bibitem{occupancy2012gaussian}
S.~T. O’Callaghan and F.~T. Ramos, ``Gaussian process occupancy maps,''
  \emph{The International Journal of Robotics Research}, 2012.

\bibitem{OccMaps}
S.~{Kim} and J.~{Kim}, ``Building occupancy maps with a mixture of gaussian
  processes,'' in \emph{IEEE International Conference on Robotics and
  Automation (ICRA)}, 2012, pp. 4756--4761.

\bibitem{occupancy2016review}
E.~G. Tsardoulias, A.~Iliakopoulou, A.~Kargakos, and L.~Petrou, ``A review of
  global path planning methods for occupancy grid maps regardless of obstacle
  density,'' \emph{Int. J. of Rob. Res.(IJRR)}, 2016.

\bibitem{OccMaps2}
S.~{Kim} and J.~{Kim}, ``Occupancy mapping and surface reconstruction using
  local {G}aussian {P}rocesses with {K}inect {S}ensors,'' in \emph{IEEE
  Transactions on Cybernetics}, 2013, pp. 1335--1346.

\bibitem{TrajOpt}
J.~Schulman, Y.~Duan, J.~Ho, A.~Lee, I.~Awwal, H.~Bradlow, J.~Pan, S.~Patil,
  K.~Goldberg, and P.~Abbeel, ``Motion planning with sequential convex
  optimization and convex collision checking,'' \emph{The International Journal
  of Robotics Research}, vol.~33, no.~9, pp. 1251--1270, 2014.

\bibitem{Helen2016signed}
H.~Oleynikova, A.~Millane, Z.~Taylor, E.~Galceran, J.~Nieto, and R.~Siegwart,
  ``Signed distance fields: A natural representation for both mapping and
  planning,'' in \emph{RSS 2016 Workshop: Geometry and Beyond-Representations,
  Physics, and Scene Understanding for Robotics}, 2016.

\bibitem{Voxblox}
H.~{Oleynikova}, Z.~{Taylor}, M.~{Fehr}, R.~{Siegwart}, and J.~{Nieto},
  ``Voxblox: Incremental 3d euclidean signed distance fields for on-board mav
  planning,'' in \emph{2017 IEEE/RSJ International Conference on Intelligent
  Robots and Systems (IROS)}, 2017, pp. 1366--1373.

\bibitem{oleynikova2018loco}
H.~Oleynikova, Z.~Taylor, R.~Siegwart, and J.~Nieto, ``Safe local exploration
  for replanning in cluttered unknown environments for micro-aerial vehicles,''
  \emph{IEEE Robotics and Automation Letters}, 2018.

\bibitem{han2019fiesta}
L.~Han, F.~Gao, B.~Zhou, and S.~Shen, ``Fiesta: Fast incremental euclidean
  distance fields for online motion planning of aerial robots,'' in \emph{2019
  IEEE/RSJ International Conference on Intelligent Robots and Systems
  (IROS)}.\hskip 1em plus 0.5em minus 0.4em\relax IEEE, 2019, pp. 4423--4430.

\bibitem{ortiz2022isdf}
J.~Ortiz, A.~Clegg, J.~Dong, E.~Sucar, D.~Novotny, M.~Zollhoefer, and
  M.~Mukadam, ``isdf: Real-time neural signed distance fields for robot
  perception,'' \emph{arXiv preprint arXiv:2204.02296}, 2022.

\bibitem{rovio}
M.~Bloesch, M.~Burri, S.~Omari, M.~Hutter, and R.~Siegwart, ``Iterated extended
  kalman filter based visual-inertial odometry using direct photometric
  feedback,'' \emph{Int. J. of Rob. Res.(IJRR)}, pp. 1053--1072, 2017.

\bibitem{gropp2020learning_shapes}
A.~Gropp, L.~Yariv, N.~Haim, M.~Atzmon, and Y.~Lipman, ``Implicit geometric
  regularization for learning shapes,'' \emph{arXiv preprint arXiv:2002.10099},
  2020.

\bibitem{park2019deepsdf}
J.~J. Park, P.~Florence, J.~Straub, R.~Newcombe, and S.~Lovegrove, ``Deepsdf:
  Learning continuous signed distance functions for shape representation,'' in
  \emph{Proceedings of the IEEE/CVF conference on computer vision and pattern
  recognition}, 2019, pp. 165--174.

\bibitem{mildenhall2021nerf}
B.~Mildenhall, P.~P. Srinivasan, M.~Tancik, J.~T. Barron, R.~Ramamoorthi, and
  R.~Ng, ``Nerf: Representing scenes as neural radiance fields for view
  synthesis,'' \emph{Communications of the ACM}, vol.~65, no.~1, pp. 99--106,
  2021.

\bibitem{mescheder2019occupancy_networks}
L.~Mescheder, M.~Oechsle, M.~Niemeyer, S.~Nowozin, and A.~Geiger, ``Occupancy
  networks: Learning 3d reconstruction in function space,'' in
  \emph{Proceedings of the IEEE/CVF conference on computer vision and pattern
  recognition}, 2019, pp. 4460--4470.

\bibitem{sitzmann2020implicit}
V.~Sitzmann, J.~Martel, A.~Bergman, D.~Lindell, and G.~Wetzstein, ``Implicit
  neural representations with periodic activation functions,'' \emph{Advances
  in Neural Information Processing Systems}, vol.~33, pp. 7462--7473, 2020.

\bibitem{sucar2021imap}
E.~Sucar, S.~Liu, J.~Ortiz, and A.~J. Davison, ``imap: Implicit mapping and
  positioning in real-time,'' in \emph{Proceedings of the IEEE/CVF
  International Conference on Computer Vision}, 2021, pp. 6229--6238.

\bibitem{rosinol2022nerf_slam}
A.~Rosinol, J.~J. Leonard, and L.~Carlone, ``Nerf-slam: Real-time dense
  monocular slam with neural radiance fields,'' \emph{arXiv preprint
  arXiv:2210.13641}, 2022.

\bibitem{zhu2022nice_slam}
Z.~Zhu, S.~Peng, V.~Larsson, W.~Xu, H.~Bao, Z.~Cui, M.~R. Oswald, and
  M.~Pollefeys, ``Nice-slam: Neural implicit scalable encoding for slam,'' in
  \emph{Proceedings of the IEEE/CVF Conference on Computer Vision and Pattern
  Recognition}, 2022, pp. 12\,786--12\,796.

\bibitem{driess2022learningSDF_planning}
D.~Driess, J.-S. Ha, M.~Toussaint, and R.~Tedrake, ``Learning models as
  functionals of signed-distance fields for manipulation planning,'' in
  \emph{Conference on Robot Learning}.\hskip 1em plus 0.5em minus 0.4em\relax
  PMLR, 2022, pp. 245--255.

\bibitem{adamkiewicz2022planning_using_NeRFs}
M.~Adamkiewicz, T.~Chen, A.~Caccavale, R.~Gardner, P.~Culbertson, J.~Bohg, and
  M.~Schwager, ``Vision-only robot navigation in a neural radiance world,''
  \emph{IEEE Robotics and Automation Letters}, vol.~7, no.~2, pp. 4606--4613,
  2022.

\bibitem{pantic2022NeRF_planning}
M.~Pantic, C.~Cadena, R.~Siegwart, and L.~Ott, ``Sampling-free obstacle
  gradients and reactive planning in neural radiance fields (nerf),''
  \emph{arXiv preprint arXiv:2205.01389}, 2022.

\bibitem{lee2019thesis}
B.~Lee, \emph{Probabilistic Online Learning of Appearance and Structure for
  Robotics}.\hskip 1em plus 0.5em minus 0.4em\relax University of Pennsylvania,
  2019.

\bibitem{Varadhan}
S.~R.~S. Varadhan, ``On the behavior of the fundamental solution of the heat
  equation with variable coefficients,'' \emph{Communications on pure and
  applied mathematics}, pp. 431--455, 1967.

\bibitem{lau2010improved}
B.~Lau, C.~Sprunk, and W.~Burgard, ``Improved updating of euclidean distance
  maps and voronoi diagrams,'' in \emph{2010 IEEE/RSJ International Conference
  on Intelligent Robots and Systems}, 2010.

\bibitem{Belyaev}
A.~G. Belyaev and P.~A. Fayolle, ``On variational and {PDE}-based distance
  function approximations,'' \emph{Computer Graphics Forum}, 2015.

\bibitem{GPbook}
C.~E. Rasmussen and C.~K. Williams, \emph{Gaussian Processes for Machine
  Learning}.\hskip 1em plus 0.5em minus 0.4em\relax Cambridge, Mass.: MIT
  Press, 2006.

\bibitem{liu2021active}
L.~Liu, S.~Fryc, L.~Wu, T.~L. Vu, G.~Paul, and T.~Vidal-Calleja, ``Active and
  interactive mapping with dynamic gaussian process implicit surfaces for
  mobile manipulators,'' \emph{IEEE Robotics and Automation Letters}, vol.~6,
  no.~2, pp. 3679--3686, 2021.

\bibitem{GPmap}
S.~Kim and J.~Kim, \emph{GPmap: A Unified Framework for Robotic Mapping Based
  on Sparse Gaussian Processes}.\hskip 1em plus 0.5em minus 0.4em\relax
  Springer International Publishing, 2015, pp. 319--332.

\bibitem{hartikainen2010kalman}
J.~Hartikainen and S.~Särkkä, ``Kalman filtering and smoothing solutions to
  temporal gaussian process regression models,'' in \emph{2010 IEEE
  International Workshop on Machine Learning for Signal Processing}, 2010, pp.
  379--384.

\bibitem{Whittle}
P.~Whittle, ``On stationary processes in the plane,'' \emph{Biometrika}, 1954.

\bibitem{paper:D-SKI}
D.~Eriksson, D.~Kun, E.~H. Lee, D.~Bindel, and A.~G. Wilson, ``Scaling gaussian
  process regression with derivatives,'' \emph{Neural Information Processing
  Systems (NIPS)}, 2018.

\bibitem{grisetti2010tutorial}
G.~Grisetti, R.~K{\"u}mmerle, C.~Stachniss, and W.~Burgard, ``A tutorial on
  graph-based slam,'' \emph{IEEE Intelligent Transportation Systems Magazine},
  vol.~2, no.~4, pp. 31--43, 2010.

\bibitem{bai2021sparse}
F.~Bai, T.~Vidal-Calleja, and G.~Grisetti, ``Sparse pose graph optimization in
  cycle space,'' \emph{IEEE Transactions on Robotics}, vol.~37, no.~5, pp.
  1381--1400, 2021.

\bibitem{points_denoise}
R.~B. Rusu, Z.~C. Marton, N.~Blodow, M.~Dolha, and M.~Beetz, ``Towards 3d point
  cloud based object maps for household environments,'' \emph{Robotics and
  Autonomous Systems}, vol.~56, no.~11, pp. 927--941, 2008.

\bibitem{paper:KISS-GP}
A.~G. Wilson and H.~Nickisch, ``Kernel interpolation for scalable structured
  gaussian processes (kiss-gp),'' \emph{International Conference on Machine
  Learning}, pp. 1775--1784, 2015.

\bibitem{paper:Kun}
K.~Dong, D.~Eriksson, H.~Nickisch, D.~Bindel, and A.~G. Wilson, ``Scalable log
  determinants for gaussian process kernel learning,'' \emph{Neural Information
  Processing Systems (NIPS)}, pp. 6330--6340, 2017.

\bibitem{paper:Quintic}
E.~H.~W. Meijering, K.~J. Zuiderveld, and M.~A. Viergever, ``Image
  reconstruction by convolution with symmetrical piecewise nth-order polynomial
  kernels,'' \emph{IEEE Transactions on Image Processing}, 1999.

\bibitem{marching}
W.~E. Lorensen and H.~E. Cline, ``Marching cubes: A high resolution 3d surface
  construction algorithm,'' in \emph{Proc. of the Conf. on Computer Graphics
  and Interactive Techniques}, 1987.

\bibitem{soorajanilkumar}
S.~Anilkumar, ``Lidar slam,''
  \url{https://github.com/soorajanilkumar/Lidar_SLAM}, 2019.

\bibitem{segal2009generalized}
A.~Segal, D.~Haehnel, and S.~Thrun, ``Generalized-icp.'' in \emph{Robotics:
  science and systems}, vol.~2, no.~4.\hskip 1em plus 0.5em minus 0.4em\relax
  Seattle, WA, 2009, p. 435.

\bibitem{kummerle2009}
R.~K{\"u}mmerle, B.~Steder, C.~Dornhege, M.~Ruhnke, G.~Grisetti, C.~Stachniss,
  and A.~Kleiner, ``On measuring the accuracy of slam algorithms,''
  \emph{Autonomous Robots}, vol.~27, no.~4, pp. 387--407, 2009.

\bibitem{scan_matching}
A.~Censi, ``Scan matching in a probabilistic framework,'' in \emph{Proc. of
  IEEE ICRA}.\hskip 1em plus 0.5em minus 0.4em\relax IEEE, 2006, pp.
  2291--2296.

\bibitem{graph_mapping}
G.~Grisetti, C.~Stachniss, S.~Grzonka, and W.~Burgard, ``A tree
  parameterization for efficiently computing maximum likelihood maps using
  gradient descent.'' in \emph{Robotics: Science and Systems}, vol.~3, 2007,
  p.~9.

\bibitem{sturm12iros_ws}
J.~Sturm, W.~Burgard, and D.~Cremers, ``Evaluating egomotion and
  structure-from-motion approaches using the {TUM RGB-D} benchmark,'' in
  \emph{Workshop on Color-Depth Camera Fusion in Robotics at the IEEE/RJS Int.
  Conf. on Intelligent Robot Systems (IROS)}, 2012.

\bibitem{masha_info_gathering}
M.~Popovi\'c, T.~Vidal-Calleja, G.~Hitz, J.~J. Chung, I.~Sa, R.~Siegwart, and
  J.~Nieto, ``An informative path planning framework for uav-based terrain
  monitoring,'' \emph{Autonomous Robots}, vol.~44, p. 889–911, 2020.

\bibitem{brian_MIUCB}
K.~M.~B. Lee, F.~Kong, R.~Cannizzaro, J.~L. Palmer, D.~Johnson, C.~Yoo, and
  R.~Fitch, ``An upper confidence bound for simultaneous exploration and
  exploitation in heterogeneous multi-robot systems,'' in \emph{Proc. of IEEE
  International Conference on Robotics and Automation (ICRA)}, 2021, pp.
  8685--8691.

\end{thebibliography}

\vspace{-40pt}

\begin{IEEEbiography}[{\includegraphics[width=1in,height=1.25in,clip,keepaspectratio]{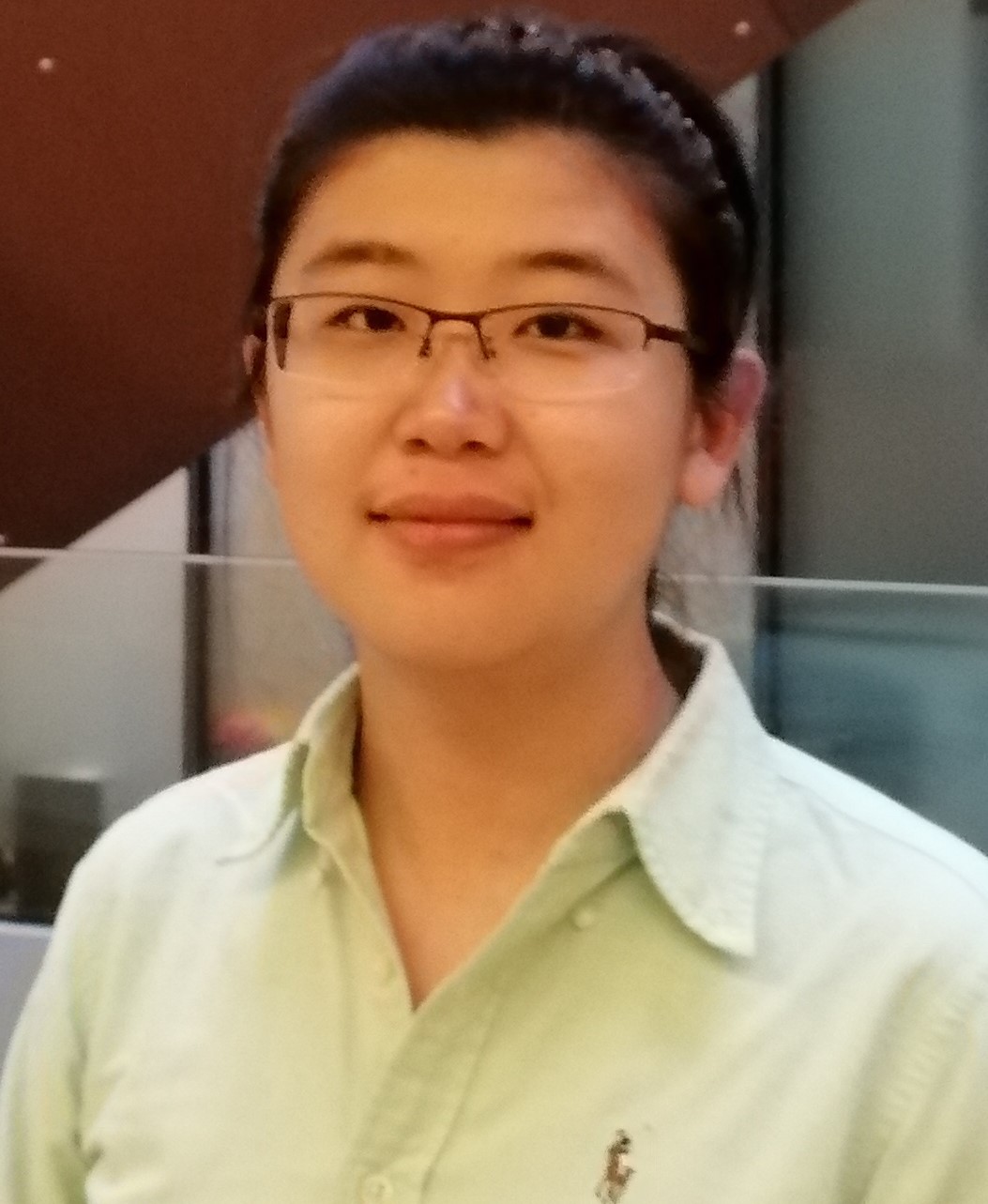}}]{Lan Wu} (Member, IEEE) received her B. Eng degree in electronic engineering in 2012, and her Ph.D. degree in robotics in 2023. Prior to joining the Robotics Institute (formerly Centre for Autonomous Systems, University of Technology Sydney) in 2019 for her Ph.D. degree, she worked in the industry as an electronic engineer, responsible for hardware and firmware designs for embedded systems. She is currently working as a postdoctoral research fellow at the Robotics Institute, University of Technology Sydney, Australia. Her research focus is probabilistic perception with depth sensors to perform efficient and effective mapping and accurate localisation for robotic systems.
\end{IEEEbiography}

\vspace{-40pt}

\begin{IEEEbiography}[{\includegraphics[width=1in,height=1.25in,clip,keepaspectratio]{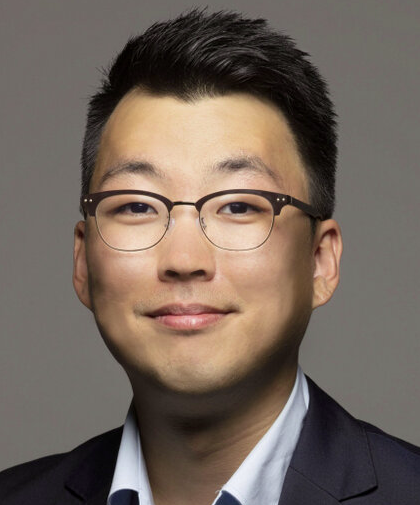}}]{Ki Myung Brian Lee} (Member, IEEE) is currently a postdoctoral research fellow at the Robotics Institute, University of Technology Sydney (UTS), Ultimo, Australia. He completed his PhD at UTS in 2023, and his BEng at the University of Sydney in 2017. He is interested in Bayesian methods for robot perception with physics- or data-driven priors, and information-aware planning enabled by such perception models. More broadly, he is interested in mobile robot autonomy in previously unseen environments. 
\end{IEEEbiography}

\vspace{-40pt}

\begin{IEEEbiography}[{\includegraphics[width=1in,height=1.25in,clip,keepaspectratio]{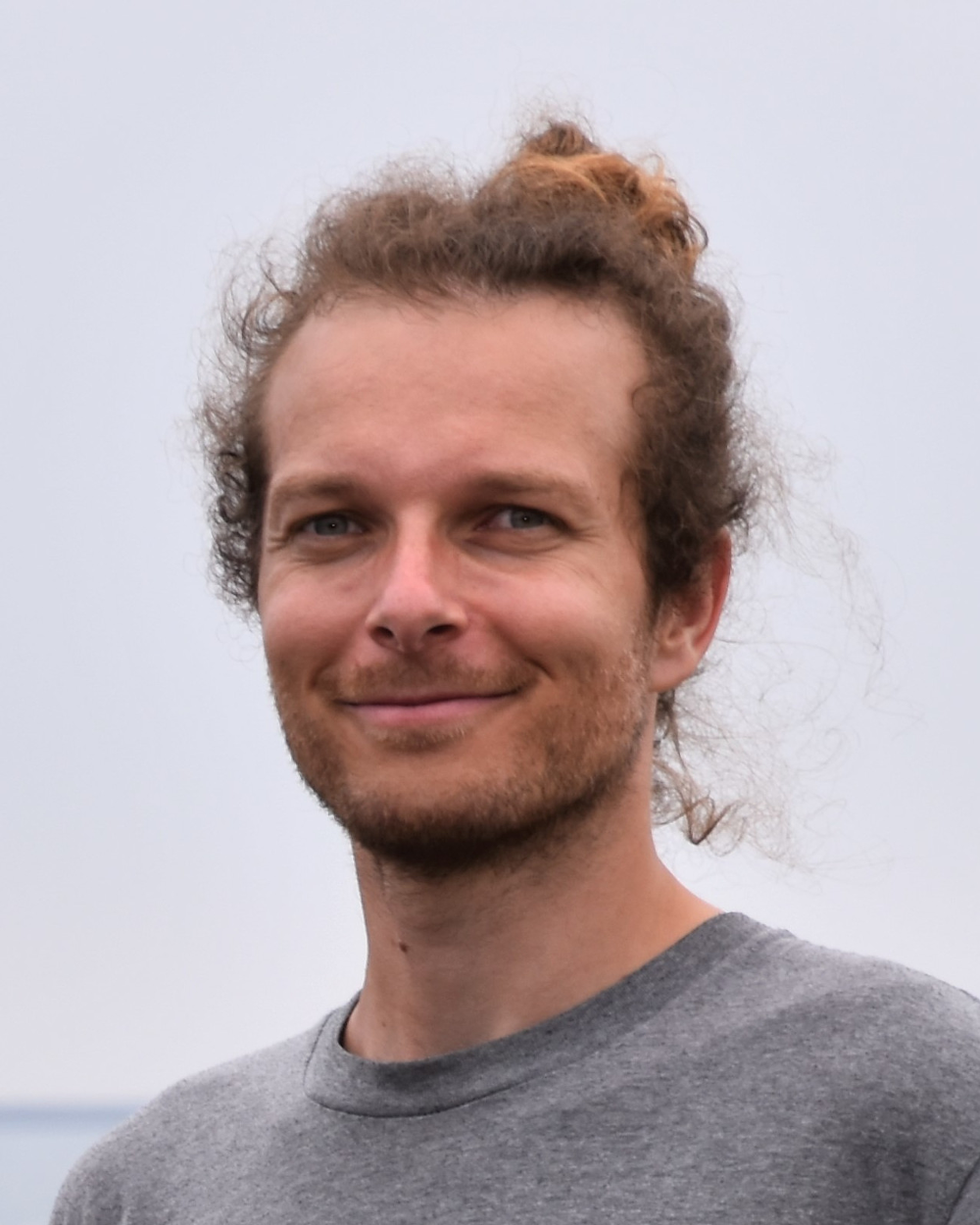}}]{Cedric Le Gentil} (Member, IEEE) received a DUT (associate degree) from Paris-Sud University (France) in 2012, and an M.Sc. from CentralSupelec (France) in 2015. He graduated with his PhD in robotics perception and state estimation in 2021 at the Robotics Institute of the University of Technology Sydney (Australia) where he is now a postdoctoral research fellow. Cedric's research interests revolve mostly around robotics perception and state estimation using various modalities: IMU, lidar, event cameras, ultrasound, etc. His past and present research includes multiple international collaborations and academic visits: the German aerospace centre (DLR) in Germany, the Autonomous Systems Lab at ETHZ in Switzerland, and the DREAM lab at GeorgiaTech Europe in France.
\end{IEEEbiography}

\vspace{-40pt}

\begin{IEEEbiography}[{\includegraphics[width=1in,height=1.25in,clip,keepaspectratio]{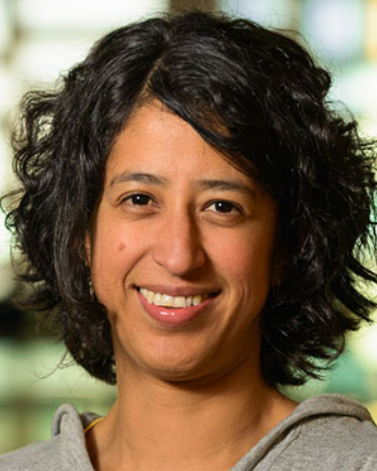}}]{Teresa Vidal-Calleja} (Senior Member, IEEE) received her BEng in mechanical engineering from the National Autonomous University of Mexico, in 2000, the MSc in mechatronics from CINVESTAV-IPN, Mexico, in 2002, and the PhD in automatic control, computer vision,
and robotics from the Polytechnic University of Catalonia, Spain, in 2007. She was a Postdoctoral Research Fellow with LAAS-CNRS, France, and the Australian Centre for Field Robotics, the University of Sydney, Australia. In 2012, she joined the Robotics Institute at the University of Technology Sydney (UTS), Australia where she was a Chancellors Research Fellow and currently is Associated Professor and Research Director. She has been a Visiting Scholar with the Active Vision Laboratory, University of Oxford, U.K., with the Autonomous Systems Lab, ETH Zürich, Switzerland, and with the Institute of Robotics and Mechatronics of the German Aerospace Centre (DLR), Germany. Her research focus is on robotic perception combining estimation theory and machine learning.
\end{IEEEbiography}

\end{document}